\documentclass[runningheads]{llncs}
\usepackage{graphicx}

\usepackage{tikz}
\usepackage{comment}
\usepackage{amsmath,amssymb} 
\usepackage{color}
\usepackage{booktabs}
\usepackage{xfrac}
\usepackage{multirow}
\usepackage{bm}
\usepackage{xcolor}
\usepackage{capt-of}
\definecolor{dodgerblue}{rgb}{0.12, 0.56, 1.0}

\usepackage[accsupp]{axessibility}  

\usepackage{xspace}
\newcommand*{\eg}{e.g.\@\xspace}

\makeatletter
\newcommand*{\etc}{%
	\@ifnextchar{.}%
	{etc}%
	{etc.\@\xspace}%
}
\makeatother

\makeatletter
\newcommand{\printfnsymbol}[1]{%
	\textsuperscript{\@fnsymbol{#1}}%
}
\makeatother

\begin{document}
	\pagestyle{headings}
	\mainmatter
	\def\ECCVSubNumber{112}  
	
	\title{On Efficient Transformer-Based Image Pre-training for Low-Level Vision
	\vspace{-0.1in}} 

	%
	%
	%
	
	\titlerunning{On Efficient Transformer-Based Image Pre-training for Low-Level Vision}
	%
	
	\author{Wenbo Li\inst{1}\thanks{These authors contributed equally.} \and Xin Lu\inst{2}\printfnsymbol{1} \and Shengju Qian\inst{1} \and Jiangbo Lu\inst{3} \and Xiangyu Zhang\inst{2} \and Jiaya Jia\inst{1}}
	\authorrunning{W. Li, et al.}
	%
	\institute{$^{1}$The Chinese University of Hong Kong  $^{2}$Megvii Technology \ $^{3}$SmartMore Technology \\
		\email{\{wenboli,sjqian,leojia\}@cse.cuhk.edu.hk} \\  \email{\{luxin,zhangxiangyu\}@megvii.com} \quad \email{jiangbo@smartmore.com}
		\vspace{-0.15in}
	}
	\maketitle
	
	\begin{abstract}
		Pre-training has marked numerous state of the arts in high-level computer vision, while few attempts have ever been made to investigate how pre-training acts in image processing systems. In this paper, we tailor transformer-based pre-training regimes that boost various low-level tasks. To comprehensively diagnose the influence of pre-training, we design a whole set of principled evaluation tools that uncover its effects on internal representations. The observations demonstrate that pre-training plays strikingly different roles in low-level tasks. For example, pre-training introduces more local information to higher layers in super-resolution (SR), yielding significant performance gains, while pre-training hardly affects internal feature representations in denoising, resulting in limited gains. Further, we explore different methods of pre-training, revealing that multi-related-task pre-training is more effective and data-efficient than other alternatives. Finally, we extend our study to varying data scales and model sizes, as well as comparisons between transformers and CNNs-based architectures. Based on the study, we successfully develop state-of-the-art models for multiple low-level tasks. Code is released at \url{https://github.com/fenglinglwb/EDT}.
		

	\end{abstract}
	\vspace{-0.25in}
	
	\section{Introduction}
	\label{sec:intro}
	
	Image pre-training has received great attention in computer vision, especially prevalent in object detection and segmentation~\cite{girshick2014rich,girshick2015fast,donahue2014decaf,long2015fully,chen2017deeplab}. When task-specific data is limited, pre-training helps models see large-scale data, thus vastly enhancing their capabilities. In the field of high-level vision, previous work~\cite{sharif2014cnn,kornblith2019better,mahajan2018exploring,sun2017revisiting,kolesnikov2020big} has shown that ConvNets pre-trained on ImageNet~\cite{deng2009imagenet} classification yield significant improvements on a wide spectrum of downstream tasks. As for image processing tasks, \eg, super-resolution (SR), denoising and deraining, the widely used datasets typically contain only a few thousand images, pointing out the potential of pre-training. However, its crucial role in low-level vision is commonly omitted. To the best of our knowledge, the sole pioneer exploring this point is IPT~\cite{chen2021pre}. Hence, there still lacks principled analysis on understanding how pre-training acts and how to perform effective pre-training.

	Previous image processing systems majorly leverage convolutional neural networks (CNNs)~\cite{lecun1989backpropagation}. More recently, transformer architectures~\cite{dosovitskiy2020image,carion2020end,liu2021Swin,wang2021pyramid,strudel2021segmenter,zhang2021multi}, initially proposed in NLP~\cite{vaswani2017attention}, have achieved promising results in vision tasks, demonstrating the potential of using transformers as a primary backbone for vision applications. Moreover, the stronger modeling capability of transformers allows for large-scale and sophisticated pre-training, which has shown great success in both NLP and computer vision~\cite{radford2018improving,radford2019language,brown2020language,devlin2018bert,raffel2020exploring,chen2021pre,he2021masked,bao2021beit,xie2021simmim}. However, it remains infeasible to directly exploit structure designs and data utilization on the \textit{full-attention} transformers for low-level vision. For example, due to the massive amount of parameters (e.g., 116M for IPT~\cite{chen2021pre}) and huge computational cost, it is prohibitively hard to explore various pre-training design choices based on IPT and further apply them in practice. Instead of following the full-attention pipeline, we explore the other \textit{window-based} variants~\cite{liang2021swinir,wang2021uformer}, which are more computationally efficient while leading to impressive performance. Along this line, we develop an encoder-decoder-based transformer (EDT) that is powerful yet efficient in data exploitation and computation. We mainly adopt EDT as a representative for efficient computation, since our observations generalize well to other frameworks, as shown in Sec.~\ref{sec:multi_task}. 
	
	In this paper, we systematically explore and evaluate how image pre-training performs in window-based transformers. Using centered kernel alignment~\cite{kornblith2019better,cortes2012algorithms} as a network ``diagnosing'' measure, we have designed a set of pre-training strategies, and thoroughly tested them with different image processing tasks. As a result, we uncover their respective effects on internal network representations, and draw useful guidelines for applying pre-training to low-level vision. The key findings and contributions of this study can be summarized as follows,
	\vspace{-0.1in}
	\begin{itemize}
		\item \textbf{Internal representations of transformers}. We find striking differences in low-level tasks. For example, SR and deraining models show clear stages, containing more local information in early layers while more global information in higher layers. The denoising model presents a relatively uniform structure filled with local information.
		\vspace{0.03in}
		\item \textbf{Effects of pre-training}. We find that pre-training improves the model performance by introducing different degrees of local information, treated as a kind of inductive bias, to the intermediate layers.
		\vspace{0.03in}
		\item \textbf{Pre-training guidelines}. Examining different pre-training strategies, we suggest a favorable \textit{multi-related-task setup} that brings more improvements and could be applied to multiple downstream tasks. Also, we find this performing strategy is more data-efficient than purely increasing the data scale. Besides, a larger model capacity usually gets more out of pre-training.
		\vspace{0.03in}
		\item \textbf{Transformers v.s. CNNs}. We observe that both transformers and CNNs benefit from pre-training, while transformers obtain greater improvements, providing higher performance baselines with fewer parameters.
		\vspace{0.03in}
		\item \textbf{SOTA models}. Based on the comprehensive study of pre-training, we provide a series of pre-trained models with state-of-the-art performance for multiple tasks, including super-resolution, denoising and deraining.
	\end{itemize}
	\vspace{-0.1in}

	\section{Encoder-Decoder-Based Transformer}
	
	Several transformers~\cite{chen2021pre,liang2021swinir,wang2021uformer} are tailored to low-level tasks, among which window-based architectures~\cite{liang2021swinir,wang2021uformer} show competitive performance under constrained parameters and computational complexity. Built upon the existing work, we make several modifications and present an efficient encoder-decoder-based transformer (EDT) in Fig.~\ref{fig:framework}. It achieves state-of-the-art results on multiple low-level tasks (see Sec.~\ref{sec:exps}), especially for those with heavy degradation. For example, EDT yields \textit{0.49}dB improvement in $\times 4$ SR on the Urban100~\cite{huang2015single} benchmark compared to IPT, while our $\times 4$ SR model size (11.6M) is only \textit{10.0\%} of IPT (115.6M) and only requires 200K images (\textit{15.6\%} of IPT) for pre-training. Also, our denoising model obtains superior performance in level-50 Gaussian denoising, with 38 GFLOPs for $192 \times 192$ inputs, far less than SwinIR~\cite{liang2021swinir} (451 GFLOPs), accounting for only \textit{8.4\%}. And the inference speed of EDT (51.9ms) is much faster than SwinIR (271.9ms). It should be pointed out that designing a novel framework is not our main purpose. Noticing similar pre-training effects on transformers in Sec.~\ref{sec:multi_task}, we adopt EDT for fast pre-training in this paper. 
	
	\subsection{Overall Architecture}
	
	\begin{figure}[t]
		\begin{center}
			\includegraphics[width=1.0\linewidth]{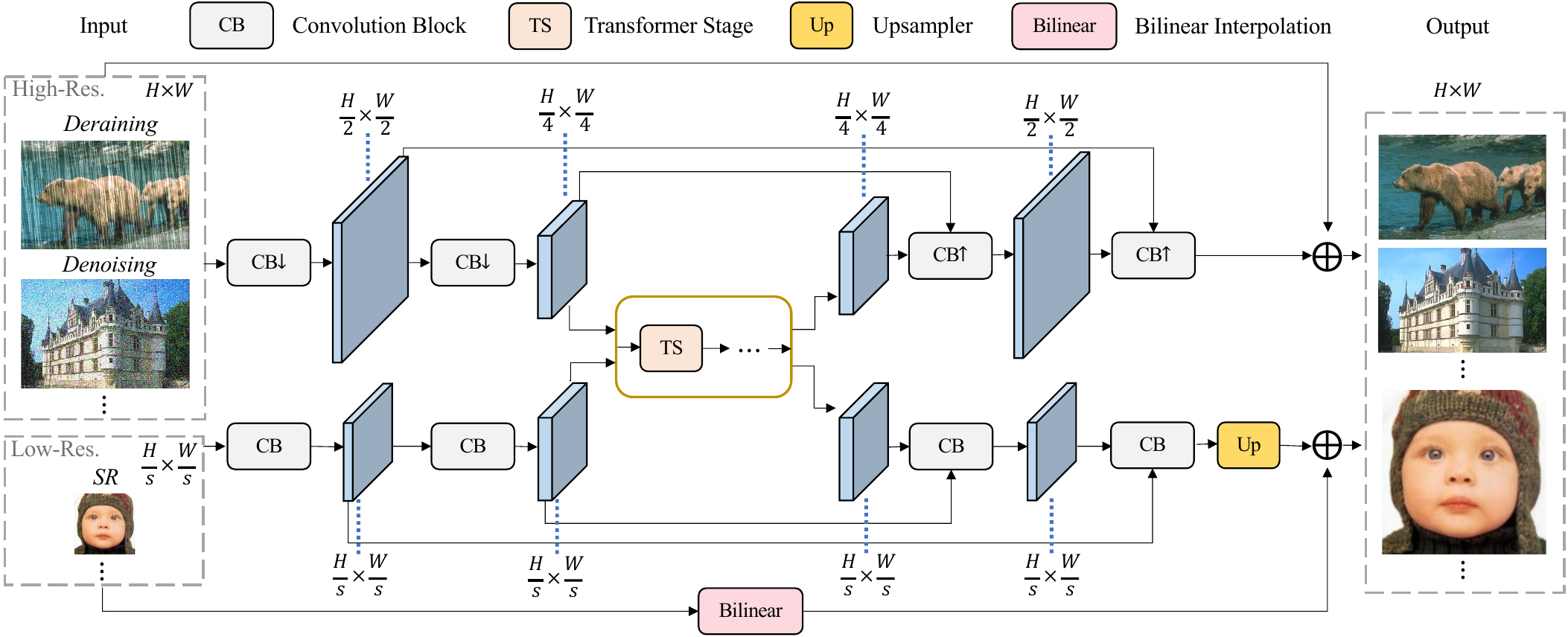}
		\end{center}
		\vspace{-0.2in}
		\caption{The proposed encoder-decoder-based transformer (EDT). EDT processes high-resolution (\eg, in deraining and denoising) and low-resolution (\eg, in SR, $s$ refers to the scale) inputs using different paths, modeling long-range interactions at a low resolution, for efficient computation.} 
		\label{fig:framework}
		\vspace{-0.1in}
	\end{figure}
	
	
	As illustrated in Fig.~\ref{fig:framework}, the proposed EDT consists of a lightweight convolution-based encoder and decoder as well as a transformer-based body, capable of efficiently modeling long-range interactions. 
	
	To improve the encoding efficiency, images are first downsampled to $\sfrac{1}{4}$ size with strided convolutions for tasks with high-resolution inputs (\eg, denoising or deraining), while being processed under the original size for those with low-resolution inputs (\eg, SR). The stack of early convolutions is also proven useful for stabling the optimization~\cite{xiao2021early}. Then, there follow multiple stages of transformer blocks, achieving a large receptive field at a low computational cost. It is noted that we improve the structure of transformer blocks through a series of ablations and provide more details in the \textit{supplementary file}. During the decoding phase, we upsample the feature back to the input size using transposed convolutions for denoising or deraining while maintaining the size for SR. Besides, skip connections are introduced to enable fast convergence during training. In particular, there is an additional convolutional upsampler before the output for super-resolution. 
	
	\subsection{Architecture Variants}
	We develop four variants of EDT with different model sizes, rendering our framework easily applied in various scenarios. As shown in Table~\ref{tab:conf}, apart from the base model (EDT-B), we also provide EDT-T (Tiny), EDT-S (Small) and EDT-L (Large). The main differences lie in the channel number, stage number and head number in the transformer body. We uniformly set the block number in each transformer stage to 6, the expansion ratio of the feed-forward network (FFN) to 2 and the window size to $(6, 24)$. 
	
	\begin{table}[t]
		\caption{Configurations of four variants of EDT. The parameter numbers and FLOPs are counted in denoising at $192 \times 192$ size.}
		\renewcommand\arraystretch{1.0}
		\begin{center}
			\begin{tabular}{|c | c | c | c | c |}
				\hline
				Models & EDT-T & EDT-S & EDT-B & EDT-L \\
				\hline
				\#Channels & 60 & 120 & 180 & 240 \\
				\#Stages & 4 & 5 & 6 & 12 \\
				\#Heads & 6 & 6 & 6 & 8 \\
				\hline
				\#Param. ($\times 10^6$, M) & 0.9 & 4.2 & 11.5 & 40.2 \\
				FLOPs ($\times 10^9$, G) & 2.8 & 12.4 & 37.6 &  136.4 \\
				\hline      
			\end{tabular}
		\end{center}
		\vspace{-0.3in}
		\label{tab:conf}
	\end{table}
	
	\section{Study of Image Pre-training}
	\label{sec:pre}
	
	\subsection{Pre-training on ImageNet}
	\label{sec:imagenet}
	
	Following~\cite{chen2021pre}, we adopt the ImageNet~\cite{deng2009imagenet} dataset in the pre-training stage. Unless specified otherwise, we only use \textit{200K} images for fast pre-training. We choose three representative low-level tasks including super-resolution (SR), denoising and deraining. Referring to~\cite{chen2021pre,agustsson2017ntire,gu2017joint}, we simulate the degradation procedure to synthesize low quality images. In terms of SR, we utilize bicubic interpolation to obtain low-resolution images. As for denoising and deraining, Gaussian noises (on RGB space) and rain streaks are directly added to the clean images. In this work, we explore $\times 2$/$\times 3$/$\times 4$ settings in SR, 15/25/50 noise levels in denoising and light/heavy rain streaks in deraining.
	
	We investigate three pre-training methods:  \textit{on a single task, on unrelated tasks} and \textit{on related tasks}. (1) The single-task pre-training refers to training a single model on a specific task (\eg, $\times 2$ SR). (2) The second is to train a single model on multiple yet unrelated tasks  (\eg, $\times 2$ SR,  level-15 denoising), while (3) the last only contains highly related tasks (\eg, $\times 2$, $\times 3$, $\times 4$ SR). As suggested in~\cite{chen2021pre}, we adopt a multi-encoder, multi-decoder, shared-body architecture for the latter two setups. The fine-tuning is performed on a single task, where the model is initialized with the pre-trained task-specific encoder and decoder as well as the shared transformer body. 
	
	\subsection{Centered Kernel Alignment}
	\label{sec:preparation}
	We introduce centered kernel alignment (CKA)\cite{kornblith2019better,cortes2012algorithms,raghu2021vision} to study representation similarity of network hidden layers, supporting quantitative comparisons within and across networks. In detail, given $m$ data points, we calculate the activations of two layers $\mathbf{X} \in \mathbb{R}^{m \times p_{1}}$ and $ \mathbf{Y} \in \mathbb{R}^{m \times p_{2}}$, having $p_{1}$ and $p_{2}$ neurons respectively. We use the Gram matrices $\mathbf{K}=\mathbf{X}\mathbf{X}^\top$ and $\mathbf{L}=\mathbf{Y}\mathbf{Y}^\top$ to compute CKA:
	\begin{equation}\label{1}
	{\rm CKA} (\mathbf{K}, \mathbf{L}) =\frac{{\rm HSIC} (\mathbf{K}, \mathbf{L})}{\sqrt{ {\rm HSIC}(\mathbf{K}, \mathbf{K}) {\rm HSIC}(\mathbf{L}, \mathbf{L})}} \,,
	\end{equation}
	where HSIC is the Hilbert-Schmidt independence criterion~\cite{gretton2007kernel}. Given the centering matrix $\mathbf{H}=\mathbf{I}_{n}-\frac{1}{n}\mathbf{1}\mathbf{1}^\top$, $\mathbf{K}^{'}=\mathbf{H}\mathbf{K}\mathbf{H}$ and $\mathbf{L}^{'}=\mathbf{H}\mathbf{L}\mathbf{H}$ are centered Gram matrices, then we have ${\rm HSIC}(\mathbf{K}, \mathbf{L}) = {\rm vec}(\mathbf{K}^{'}) \cdot {\rm vec}(\mathbf{L}^{'})/(m-1)^{2}$. Thanks to the properties of CKA, invariant to orthogonal transformation and isotropic scaling, we are able to conduct a meaningful analysis of neural network representations.  However, naive computation of CKA requires maintaining the activations across the entire dataset in memory, causing much memory consumption. To avoid this, we use minibatch estimators of CKA\cite{nguyen2020wide}, with a minibatch of 300 by iterating over the test dataset 10 times.
	
	\subsection{Representation Structure of EDT}
	\label{sec:finetune}
	
	We begin our investigation by studying the internal representation structure of our models. How are representations propagated within models in different low-level tasks?  To answer this intriguing question, we compute CKA similarities between every pair of layers within a model. Apart from the convolutional head and tail, we include outputs of attention and FFN after residual connections in the transformer body.
	
\begin{figure}[t]
		\begin{center}
			\includegraphics[width=1.0\linewidth]{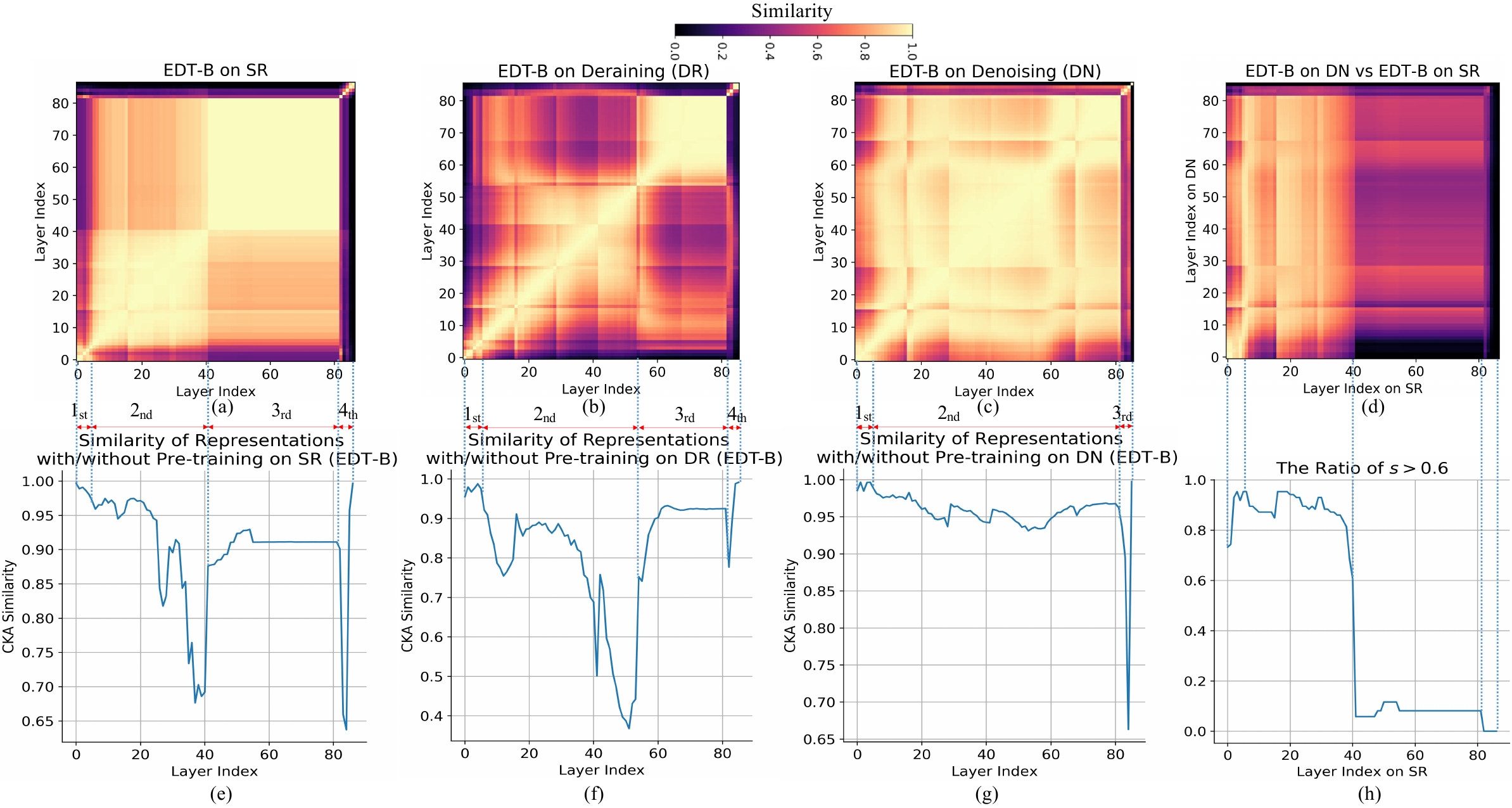}
		\end{center}
		\vspace{-0.2in}
		\caption{Sub-figures (a)-(c) show CKA similarities between all pairs of layers in $\times 2$ SR, light streak deraining and level-15 denoising EDT-B models with single-task pre-training, and the corresponding similarities between \textit{with} and \textit{without} pre-training are shown in (e)-(g). Sub-figure (d) shows the cross-model comparison between SR and denoising models and (h) shows the ratios of layer similarity larger than 0.6, where ``$s$'' means the similarity between the current layer in SR and any layer in denoising.}
		\label{fig:single}
		\vspace{-0.1in}
	\end{figure}

	We observe a block-diagonal structure in the CKA similarity maps in Fig.~\ref{fig:single}. As for the SR and deraining models in Fig.~\ref{fig:single} (a)-(b), we find there are roughly four groups, among which a range of transformer layers are of high similarity. The first and last group structures (from left to right) correspond to the model head and tail, while the second and third group structures account for the transformer body. As for the denoising task (Fig.~\ref{fig:single} (c)), there are only three obvious group structures, where the second one (transformer body) is dominated. Finally, from the cross-model comparison in Fig.~\ref{fig:single} (d) and (h), we find \textit{higher similarity} scores between denoising body layers and the second group SR layers, while showing \textit{significant differences} compared to the third group SR layers.
	
	Further, we explore the impact of single-task pre-training on the internal representations. As for SR and deraining in Fig.~\ref{fig:single} (e)-(f), the representations of the model head and tail remain basically unchanged. Meanwhile, we observe \textit{obvious representation changes} in the transition regions between the second and third groups. In terms of denoising, as shown in Fig.~\ref{fig:single} (g), the internal representations do not change too much, consistent with the finding in Table~\ref{tab:denoise} that denoising tasks obtain fewer improvements, compared to SR and deraining tasks.
	
	\vspace{0.05in}
	\noindent\textbf{Key Findings}: (1) SR and deraining models show clear stages in the internal representations of the transformer body, while the denoising model presents a relatively uniform structure; (2) the denoising model layers show more similarity to the lower layers of SR models, containing more local information, as verified in Sec.~\ref{sec:multi_task}; (3) single-task pre-training mainly affects the higher layers of SR and deraining models but has limited impact on the denoising model. 
	
	\subsection{Single- and Multi-Task Pre-training}
	\label{sec:multi_task}
	
	\begin{figure}[t]
		\begin{center}
			\includegraphics[width=1.0\linewidth]{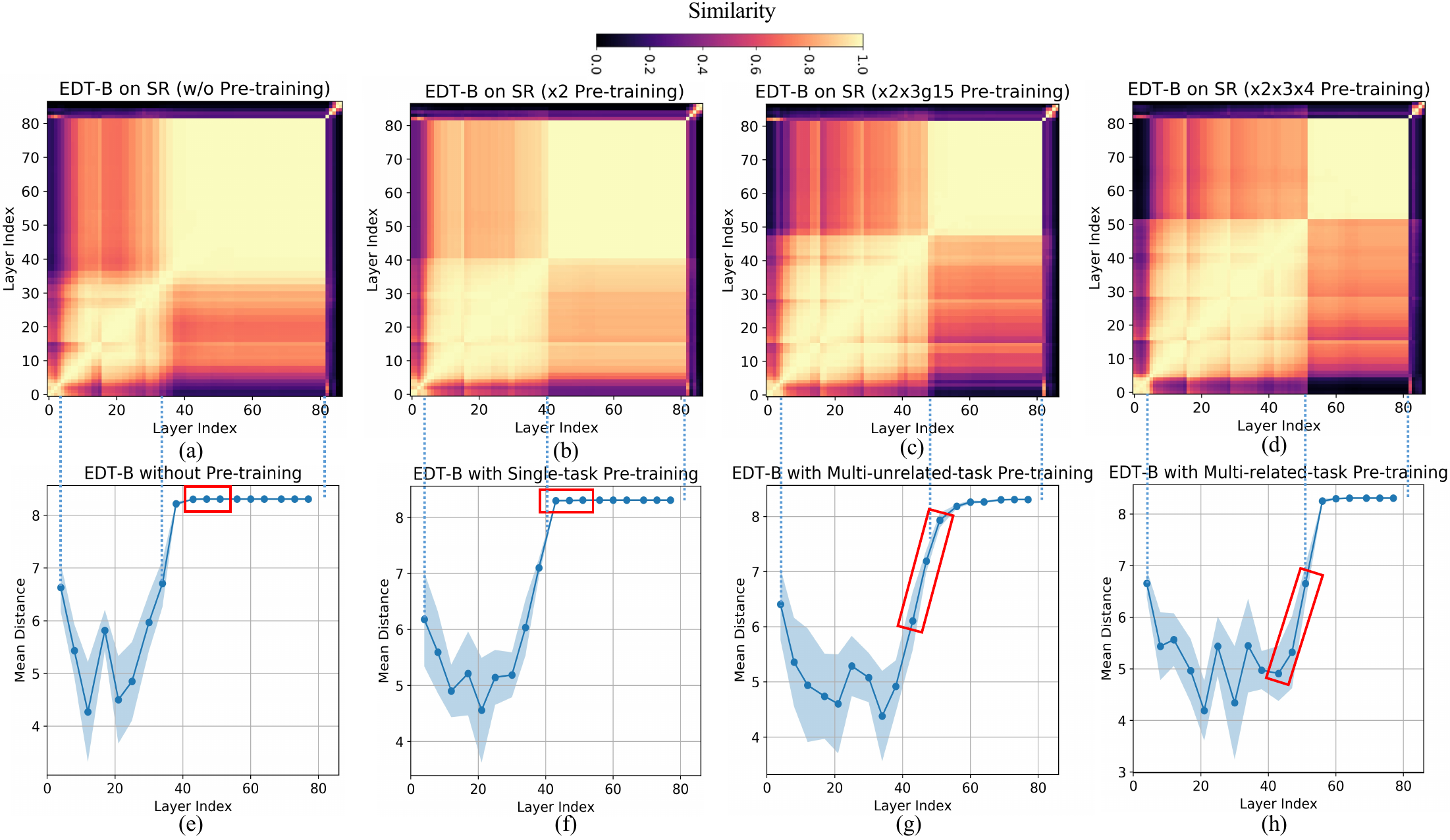}
		\end{center}
		\vspace{-0.15in}
		\caption{Sub-figures (a)-(d) show CKA similarities of $\times 2$ SR models, without pre-training as well as with pre-training on \textit{a single task} ($\times 2$), \textit{unrelated tasks} ($\times 2$, $\times 3$ SR, g15 denoising) and \textit{highly related tasks} ($\times 2$, $\times 3$, $\times 4$ SR). Sub-figures (e)-(h) show the corresponding attention head mean distances of transformer blocks. We do not plot shifted local windows in (e)-(h) so that the last blue dotted line (``\textcolor{dodgerblue}{-{}-{}-}'') has no matching point. The red boxes indicate the same attention modules.}
		\label{fig:multiple}
		\vspace{-0.1in}
	\end{figure}
	
	In the previous section, we observe that the transformer body of SR models is clearly composed of two group structures and pre-training mainly changes the representations of higher layers. What is the difference between these two partitions? How does the pre-training, especially multi-task pre-training, affect the behaviors of models?
	
	We conjecture that one possible reason causing the partition lies with the difference of ability to incorporate local or global information between different layers. We start by analyzing self-attention layers for their mechanism of dynamically aggregating information from other spatial locations, which is quite different from the fixed receptive field of the FFN layer. To represent the range of attentive fields, we average pixel distances between the queries and keys using attention weights for each head over 170,000 data points, where a larger distance usually refers to using more global information. Note that we do not record attention distances of shifted local windows, because the shift operation narrows down the boundary windows and hence can not reflect real attention distances.
	
	As shown in Fig.~\ref{fig:multiple} (e)-(h), for the second group structure (counted from the head, same as Sec.~\ref{sec:finetune}), the standard deviation of attention distances (shown as the blue area) is large and the mean value is small, indicating the attention modules in this group structure area have a mix of local heads (relatively small distances) and global heads (relatively large distances). On the contrary, the third group structure only contains global heads, showing more global information are aggregated in this stage.
	
	Compared to single-task pre-training ($\times 2$ SR, Fig.~\ref{fig:multiple} (b) and (f)), multi-unrelated-task setup ($\times 2$, $\times 3$ SR, g15 denoising, in Fig.~\ref{fig:multiple} (c) and (g)) converts more global representations (in red box) of the third group to local ones, increasing the scope of the second group. In consequence, as shown in Fig.~\ref{fig:bar}, we observe obvious PSNR improvements on all benchmarks. When replacing the g15 denoising with highly related $\times 4$ SR ($\times 2$, $\times 3$, $\times 4$ SR, in Fig.~\ref{fig:multiple} (d) and (h)), we observe more changes in global representations, along with further improvements in Fig.~\ref{fig:bar}. The inferiority of multi-unrelated-task setup is mainly due to the representation mismatch of unrelated tasks, as shown in Sec.~\ref{sec:finetune}. We also provide detailed quantitative comparisons for all tasks in the \textit{supplementary material}.
	
	
	
	\vspace{0.05in}
	\noindent\textbf{Key Findings}: (1) the representations of SR models contain more local information in early layers while more global information in higher layers; (2) all three pre-training methods can greatly improve the performance by introducing different degrees of local information, treated as a kind of inductive bias, to the intermediate layers of the model, among which \textit{multi-related-task pre-training performs best}.
	
	\begin{table}[t]
		\begin{minipage}[c]{0.49\linewidth}
			\centering
			\includegraphics[width=1.0\textwidth]{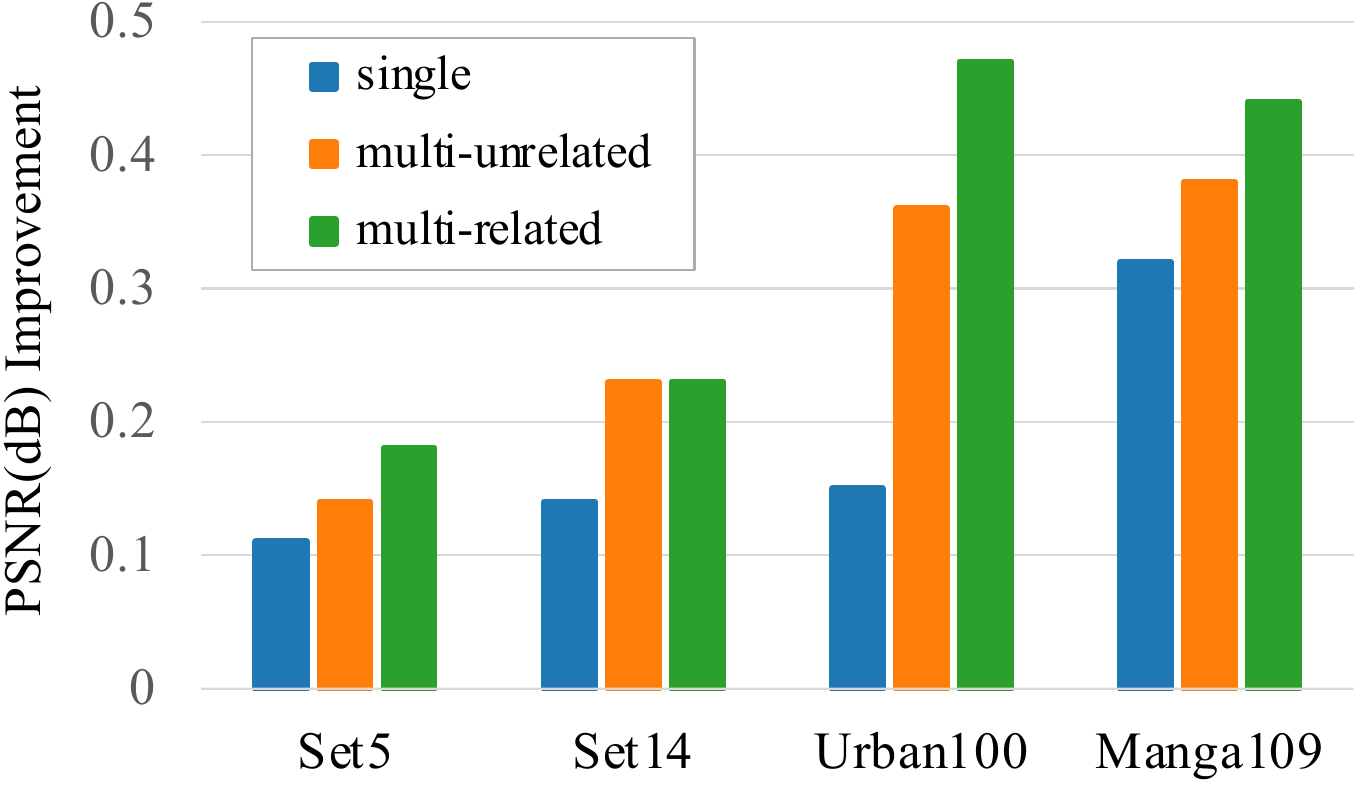}
			\captionof{figure}{PSNR improvements of single-task, multi-unrelated-task and multi-related-task pre-training for EDT-B in $\times 2$ SR.}
			\label{fig:bar}
		\end{minipage}\hfill
		\begin{minipage}[c]{0.48\linewidth}
			\centering
			\includegraphics[width=1.0\textwidth]{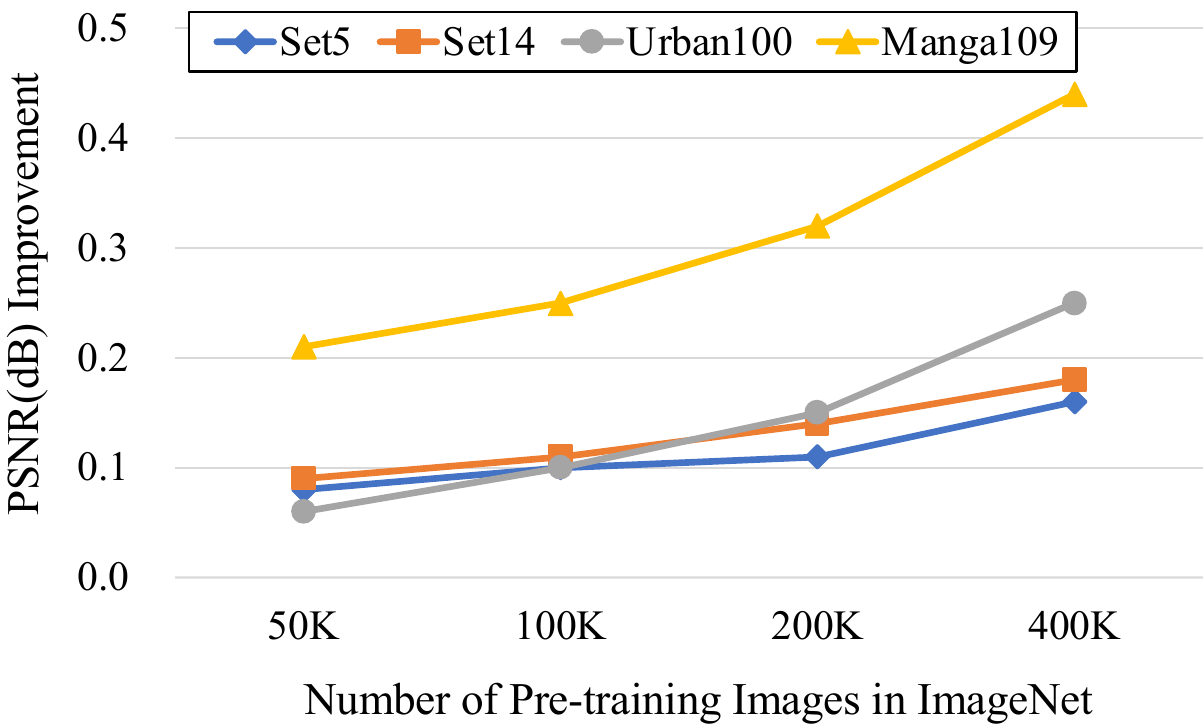}
			\captionof{figure}{PSNR improvements of different data scales during single-task pre-training for EDT-B in $\times 2$ SR.}
			\label{fig:data}
		\end{minipage}
		\vspace{-0.15in}
	\end{table}
	
\begin{figure}[t]
		\begin{center}
			\includegraphics[width=1.0\linewidth]{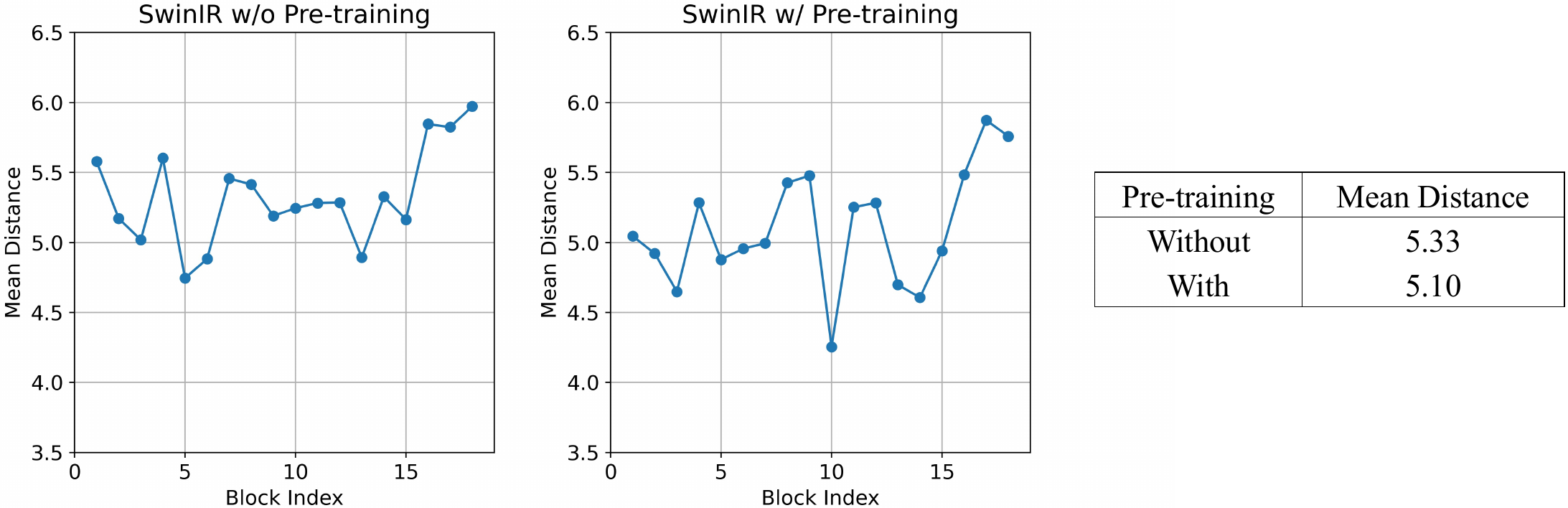}
		\end{center}
		\vspace{-0.2in}
		\caption{Attention head mean distances of transformer blocks in SwinIR~\cite{liang2021swinir} with and without pre-training.}
		\label{fig:swinir}
		\vspace{-0.15in}
	\end{figure}

	To validate whether the finding that pre-training brings more local information to the model also fit other window-based frameworks, we show the attention head distances of SwinIR~\cite{liang2021swinir} in Fig.~\ref{fig:swinir}. Without pre-training, the first few blocks (1-15) tend to be local while the last ones (16-18) are more global. And pre-training brings more local representations, matching our observation before.
	
	\subsection{Effect of Data Scale on Pre-training}
	\label{sec:scale}
	

	\begin{table}[t]
		\caption{PSNR(dB) results of different pre-training (single-task) data scales in $\times 2$ SR. ``EDT-B$^\dagger$'' refers to the base model with single-task ($\times 2$ SR) pre-training and ``EDT-B$^\star$'' represents the base model with multi-related-task ($\times 2$, $\times 3$, $\times 4$ SR) pre-training. The best results are in \textbf{bold}.}
		\setlength\tabcolsep{5pt}
		\begin{center}
			\begin{tabular}{| c | c | c  c  c  c |}
				\hline
				Model & Data & Set5 & Set14 & Urban100 & Manga109 \\
				\hline
				EDT-B & 0 & 38.45 & 34.57 & 33.80 & 39.93 \\
				EDT-B$^\dagger$ & 50K & 38.53 & 34.66 & 33.86 & 40.14 \\
				EDT-B$^\dagger$ & 100K & 38.55 & 34.68 & 33.90 & 40.18  \\
				EDT-B$^\dagger$ & 200K & 38.56 & 34.71 & 33.95 & 40.25 \\
				EDT-B$^\dagger$ & 400K & 38.61 & 34.75 & 34.05 & \textbf{40.37} \\			
				\hline
				EDT-B$^\star$ & 200K & \textbf{38.63} & \textbf{34.80} & \textbf{34.27} & \textbf{40.37} \\
				\hline
			\end{tabular}
		\end{center}
		\vspace{-0.1in}
		\label{tab:data}
	\end{table} 
	
	In this section, we investigate how pre-training data scale affects the super-resolution performance. As shown in Fig.~\ref{fig:data} and Table~\ref{tab:data}, with regard to the EDT-B model, we obviously observe incremental PSNR improvements on multiple SR benchmarks by increasing the data scale from 50K to 400K during single-task pre-training. It is noted that we double the pre-training iterations for the data scale of 400K so that the data can be fully functional. However, a longer pre-training period largely increases the training burden. 
	
	On the contrary, as shown in Table~\ref{tab:data}, multi-related-task pre-training (with much fewer training iterations) successfully breaks through the limit. Our EDT-B model with multi-related-task pre-training on 200K images achieves new state of the arts on all benchmarks, though a smaller data scale is adopted, revealing that simply increasing the data scale may not be the optimal option. Thus, we suggest multi-related-task pre-training is more effective and data-efficient in low-level vision.
	
	
	\subsection{Effect of Model Size on Pre-training}
	
	\begin{table}[t]
		\begin{minipage}[c]{0.49\linewidth}
			\centering
			\includegraphics[width=1.0\textwidth]{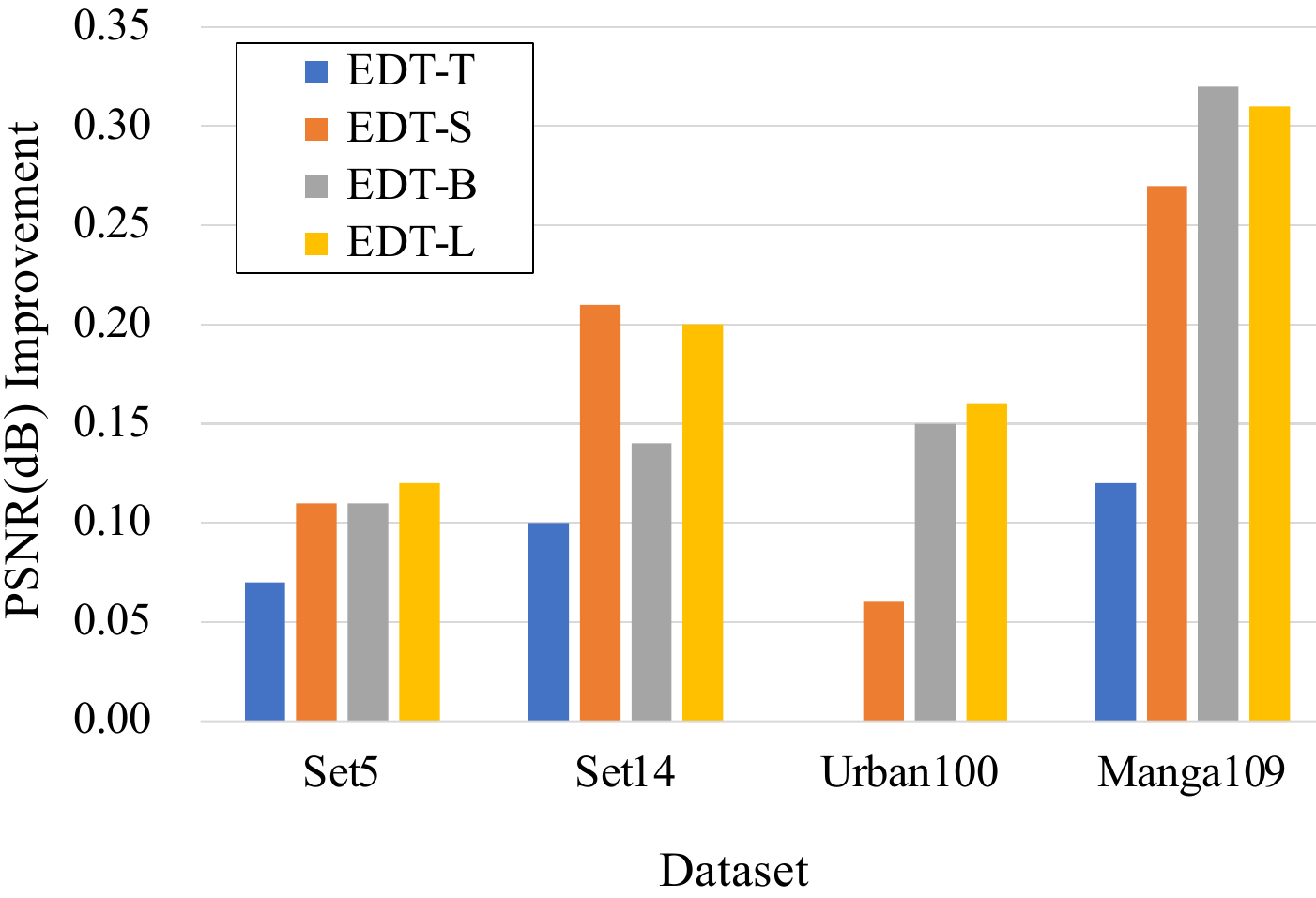}
			\captionof{figure}{PSNR improvements of four variants of EDT models using single-task pre-training in $\times 2$ SR. ``T'', ``S'', ``B'' and ``L'' refer to tiny, small, base and large models. The improvement of EDT-T on Urban100 is 0.00dB, thus we do not plot the bar.}
			\label{fig:model}
		\end{minipage}\hfill
		\begin{minipage}[c]{0.49\linewidth}
			\centering
			\caption{PSNR(dB) results of four variants of EDT models using single-task pre-training in $\times 2$ SR. ``$\dagger$'' indicates models with single-task pre-training.}
			\renewcommand\arraystretch{1.2}
			\begin{tabular}{| c | c c c c |}
				\hline
				Model & Set5 & Set14 & Urban100 & Manga109 \\
				\hline
				EDT-T & 38.23 & 33.99 & \textbf{32.98} & 39.45 \\
				EDT-T$^\dagger$ & \textbf{38.30} & \textbf{34.09} & \textbf{32.98} & \textbf{39.57} \\		
				\hline
				EDT-S & 38.38 & 34.36 & 33.61 & 39.78 \\
				EDT-S$^\dagger$ & \textbf{38.49} & \textbf{34.57} & \textbf{33.67} & \textbf{40.05} \\
				\hline
				EDT-B & 38.45 & 34.57 & 33.80 & 39.93 \\
				EDT-B$^\dagger$ & \textbf{38.56} & \textbf{34.71} & \textbf{33.95} & \textbf{40.25 }\\
				\hline
				EDT-L & 38.47 & 34.51 & 33.91 & 40.02 \\
				EDT-L$^\dagger$ & \textbf{38.59} & \textbf{34.71} & \textbf{34.07} & \textbf{40.33} \\
				\hline
			\end{tabular}
			\vspace{0.05in}
			\label{tab:model}
		\end{minipage}
		\vspace{-0.2in}
	\end{table}

	\begin{figure}[t]
		\begin{center}
			\includegraphics[width=0.9\linewidth]{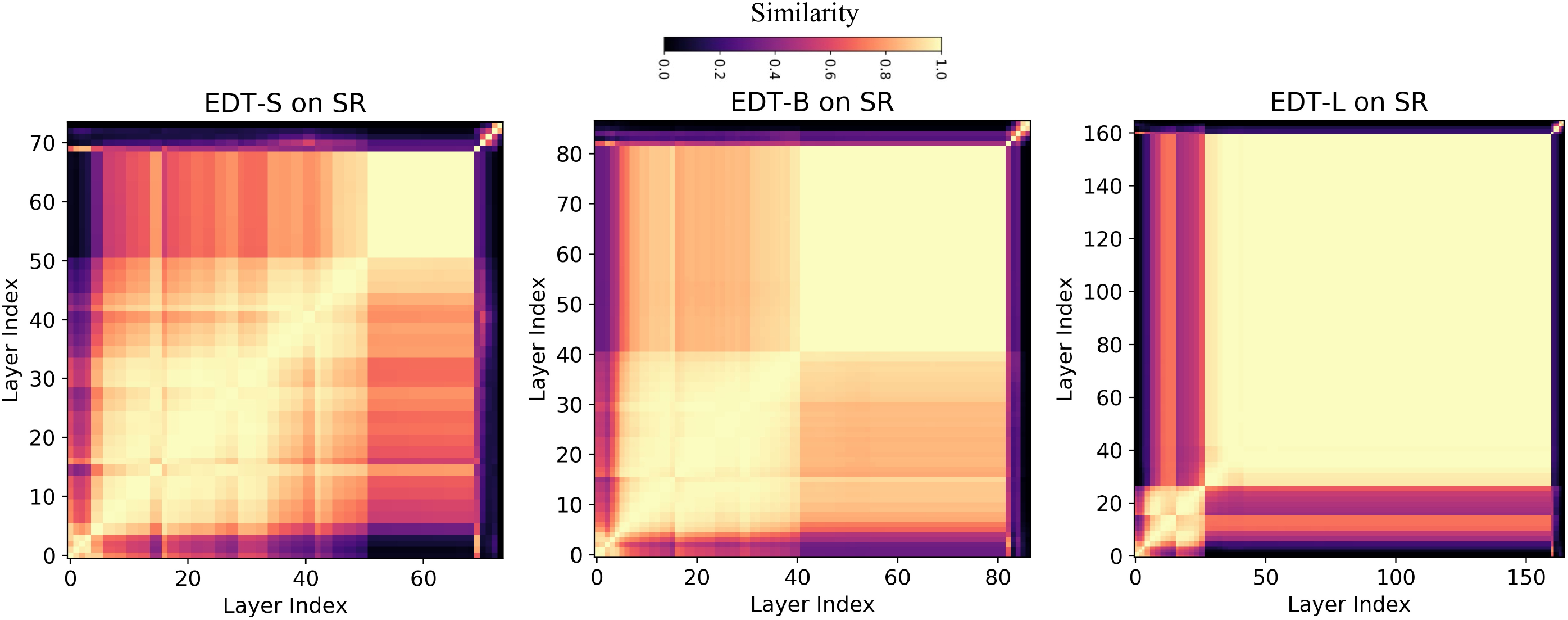}
		\end{center}
		\vspace{-0.15in}
		\caption{CKA similarities between all pairs of layers in EDT-S, EDT-B and EDT-L models using single-task pre-training in $\times 2$ SR.}
		\label{fig:edtl}
	\end{figure}
	
	We conduct experiments to compare the performance of single-task pre-training for four model variants in the $\times 2$ SR task. As shown in Fig.~\ref{fig:model}, we visualize the PSNR improvements of models with pre-training over counterparts trained from scratch. It is observed that models with larger capacities generally obtain more improvements. Especially, we find pre-training can still improve a lot upon already strong EDT-L models, showing the potential of pre-training in low-level tasks. The detailed quantitative results are provided in Table~\ref{tab:model}.
	
	Here we visualize the CKA maps of the EDT-S, EDT-B and EDT-L models in Fig~\ref{fig:edtl}. As illustrated in Sec.~\ref{sec:finetune}, we already know there are roughly four group structures in the CKA maps of SR models, among which the second and third group structures account for the transformer body. The proportion of the third part is positively correlated with the model size. Especially, compared to the other two, the third group structure of EDT-L account for the vast majority and show high similarities, which somewhat reflects the redundancy of the model.
	
	
	\begin{figure}[t]
		\begin{center}
			\includegraphics[width=0.85\linewidth]{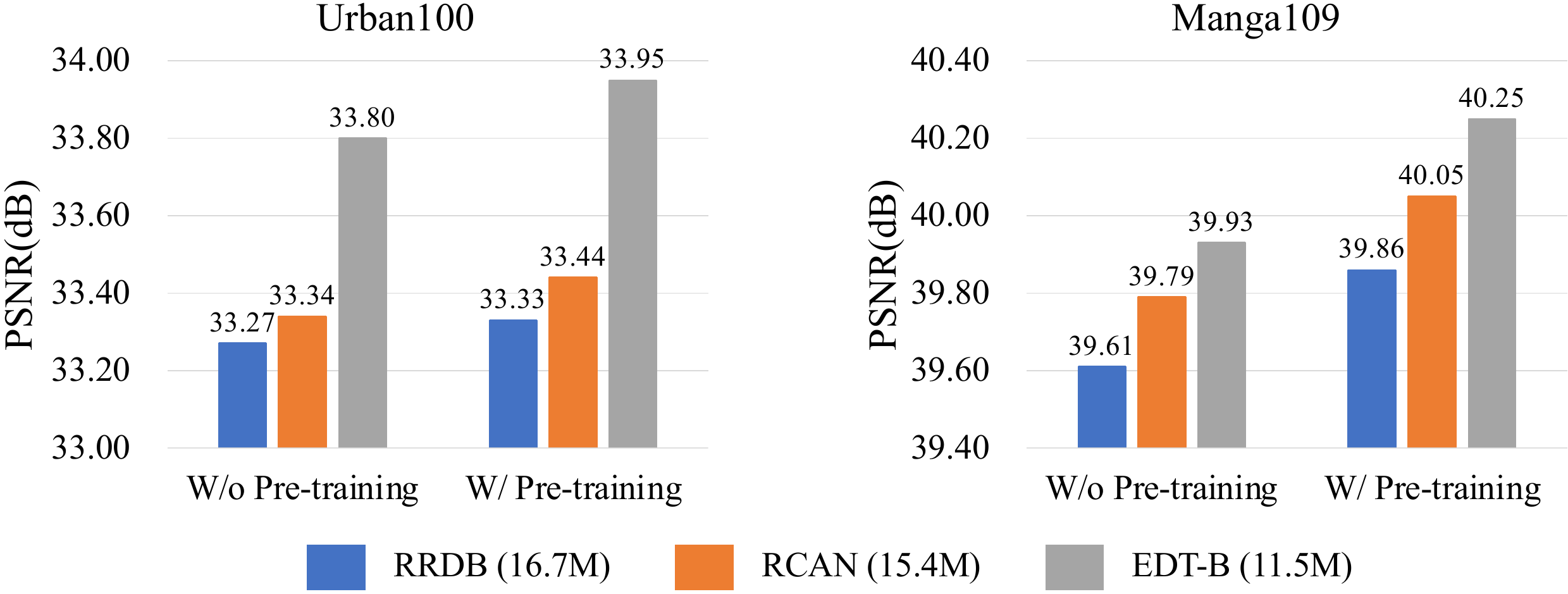}
		\end{center}
		\vspace{-0.1in}
		\caption{Quantitative comparison between ConvNets (RRDB~\cite{wang2018esrgan} and RCAN~\cite{zhang2018image}) and our EDT-B without (``W/o'') and with (``W/'') single-task pre-training in $\times 2$ SR.}
		\label{fig:convnet}
		\vspace{-0.1in}
	\end{figure}
	
	\subsection{EDT v.s. ConvNets with Pre-training}
	
	
	We further explore the relationship of internal representations between EDT and CNNs-based models (RRDB~\cite{wang2018esrgan} and RCAN~\cite{zhang2018image}) and the superiority of our transformer architecture over ConvNets. As for RRDB and RCAN, apart from the head and tail, we use outputs of blocks, \eg, residual dense blocks in RRDB
	
	\begin{table}[h]
		\begin{minipage}[c]{0.3\linewidth}
			\vspace{-0.1in}
			and residual channel attention blocks in RCAN, to compute CKA similarities.
			
			\setlength{\parindent}{2em}  As shown in Fig.~\ref{fig:cka_conv}, the early layers in EDT-B have more representations similar to those learned in RRDB and RCAN, which tend to be local as mentioned in Sec.~\ref{sec:multi_task}, while higher layers in EDT-B incorporate more global information, showing clear differences compared to ConvNets.  Besides, Fig.~\ref{fig:convnet} demonstrates that our EDT-B obtains greater or comparable improvements from pre-training, giving higher baselines with fewer parameters. We suggest that the superiority of transformers may come from the utilization of global information.

		\end{minipage}\hfill
		\begin{minipage}[c]{0.65\linewidth}
			\centering
			\includegraphics[width=1.0\textwidth]{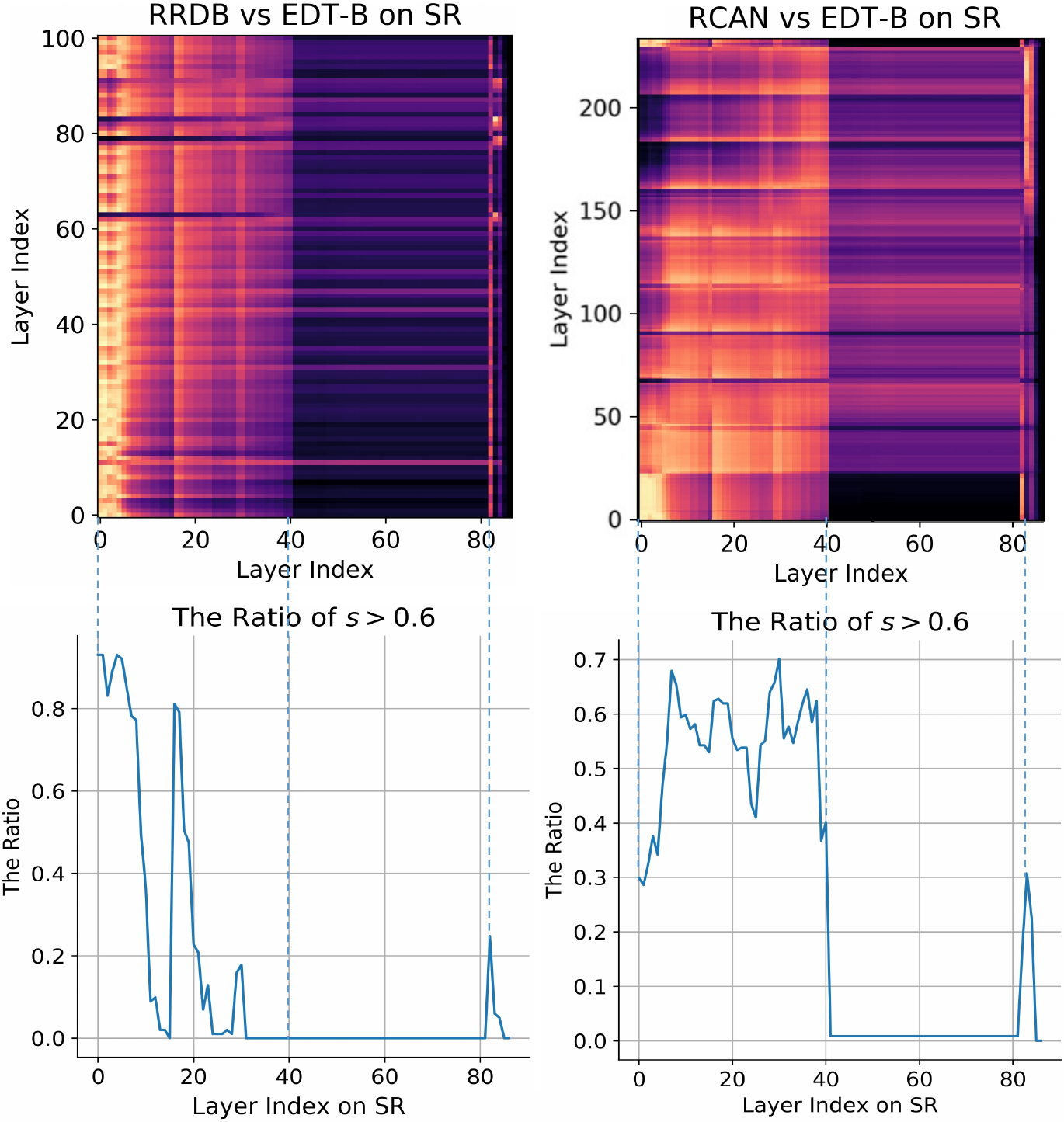}
			\captionof{figure}{The first row shows CKA similarities between ConvNets (RRDB~\cite{wang2018esrgan} and RCAN~\cite{zhang2018image}) and our EDT-B with single-task pre-training in $\times 2$ SR. The second row shows the ratios of layer similarity larger than 0.6.} 
			\label{fig:cka_conv}
		\end{minipage}
		\vspace{-0.1in}
	\end{table}
	
	\noindent

	%

	
	\section{Experiments}
	\label{sec:exps}
	Following the pre-training guidelines, we conduct experiments in super-resolution (SR), denoising and detraining. As aforementioned, we observe that multi-related-task pre-training is highly effective and data-efficient. Thus, we adopt this pre-training strategy in all the rest tests. The involved pre-training tasks of SR include $\times 2$, $\times 3$ and $\times 4$, those of denoising include g15, g25 and g50, and those of deraining include light and heavy rain streaks. More detailed experimental settings and visual comparisons are given in the \textit{supplementary material}.  
	
	\vspace{-0.1in}
	\subsection{Super-Resolution Results}
	
	For the super-resolution (SR) task, we test our models on two settings, classical and lightweight SR, where the latter generally refers to models with $<$ 1M parameters.
	
	\vspace{0.05in}
	\noindent{\textbf{Classical SR.}} We compare our EDT with state-of-the-art CNNs-based methods as well as transformer-based methods. As shown in Table~\ref{tab:classical_sr}, while the proposed EDT-B serves as a strong baseline, achieving nearly 0.1dB gains on multiple datasets over SwinIR~\cite{liang2021swinir}, pre-training still brings significant improvements on $\times 2$, $\times 3$ and $\times 4$ scales. For example, we observe up to \textit{0.46}dB and \textit{0.45}dB improvements on high-resolution benchmark Urban100~\cite{huang2015single} and Manga109~\cite{matsui2017sketch}, manifesting the effectiveness of our pre-training strategy. 
	
	\vspace{0.05in}
	\noindent{\textbf{Lightweight SR.}} Among lightweight SR algorithms in Table~\ref{tab:lightweight_sr}, our model achieves \textit{the best} results on all benchmark datasets. Though SwinIR uses larger training patches ($64 \times 64$) than ours ($48 \times 48$) and adopts the well-trained $\times 2$ model as initialization for $\times 3$ and $\times 4$ scales, our EDT-T still obtains nearly 0.2dB and 0.4dB improvements on Urban100 and Manga109 across all scales, demonstrating the superiority of our architecture.

	\begin{table}[t]
		\caption{Quantitative comparison for classical SR on PSNR(dB)/SSIM on the Y channel from the YCbCr space. ``$\ddagger$'' means the $\times 3$ and $\times 4$ models of SwinIR are pre-trained on the $\times 2$ setup and training patch size is $64 \times 64$ (ours is $48 \times 48$). ``$\dagger$'' indicates methods with a pre-training. \textcolor{red}{Best} and \textcolor{blue}{second best} results are in red and blue colors.}
		\vspace{-0.1in}
		\begin{center}
			\resizebox{\textwidth}{!}{
				\begin{tabular}{| c | l | c | c | c | c | c | c | c | c | c | c | c |}
					\hline
					\multirow{2}{*}{Scale} & \multirow{2}{*}{Method} & \#Param. & \multicolumn{2}{|c|}{Set5} & \multicolumn{2}{|c|}{Set14} & \multicolumn{2}{|c|}{BSDS100} & \multicolumn{2}{|c|}{Urban100} & \multicolumn{2}{|c|}{Manga109} \\
					\cline{4-13}
					~ & ~ & ($\times 10^6$) & PSNR & SSIM & PSNR & SSIM & PSNR & SSIM & PSNR & SSIM & PSNR & SSIM \\
					\hline
					\multirow{9}{*}{$\times 2$} & RCAN~\cite{zhang2018image} & 15.4 & 38.27 & 0.9614 & 34.12 & 0.9216 & 32.41 & 0.9027 & 33.34 & 0.9384 & 39.44 & 0.9786 \\
					~ & SAN~\cite{dai2019second} & 15.7 & 38.31 & 0.9620 & 34.07 & 0.9213 & 32.42 & 0.9028 & 33.10 & 0.9370 & 39.32 & 0.9792 \\
					~ & HAN~\cite{niu2020single} & 15.9 & 38.27 & 0.9614 & 34.16 & 0.9217 & 32.41 & 0.9027 & 33.35 & 0.9385 & 39.46 & 0.9785 \\
					~ & NLSA~\cite{mei2021image} & 31.9 & 38.34 & 0.9618 & 34.08 & 0.9231 & 32.43 & 0.9027 & 33.42 & 0.9394 & 39.59 & 0.9789 \\
					~ & IPT$^\dagger$~\cite{chen2021pre} & 115.5 & 38.37 & - & 34.43 & - & 32.48 & - & 33.76 & - & - & - \\
					~ & SwinIR~\cite{liang2021swinir} & 11.8 & 38.42 & 0.9622 & 34.48 & 0.9252 & 32.50 & 0.9038 & 33.70 & 0.9418 & 39.81 & 0.9796 \\
					~ & SwinIR{$^\ddagger$}~\cite{liang2021swinir} & 11.8 & 38.42 & 0.9623 & 34.46 & 0.9250 & \textcolor{blue}{32.53} & \textcolor{blue}{0.9041} & \textcolor{blue}{33.81} & \textcolor{blue}{0.9427} & 39.92 & 0.9797 \\
					~ & EDT-B(Ours) & 11.5 & \textcolor{blue}{38.45} & \textcolor{blue}{0.9624} & \textcolor{blue}{34.57} & \textcolor{blue}{0.9258} & 32.52 & \textcolor{blue}{0.9041} & 33.80 & 0.9425 & \textcolor{blue}{39.93} & \textcolor{blue}{0.9800} \\
					~ & \textbf{EDT-B}$^\dagger$(Ours) & 11.5 & \textcolor{red}{38.63} & \textcolor{red}{0.9632} & \textcolor{red}{34.80} & \textcolor{red}{0.9273} & \textcolor{red}{32.62} & \textcolor{red}{0.9052} & \textcolor{red}{34.27} & \textcolor{red}{0.9456} & \textcolor{red}{40.37} & \textcolor{red}{0.9811} \\
					\hline  
					
					\multirow{9}{*}{$\times 3$} & RCAN~\cite{zhang2018image} & 15.6 & 34.74 & 0.9299 & 30.65 & 0.8482 & 29.32 & 0.8111 & 29.09 & 0.8702 & 34.44 & 0.9499 \\
					~ & SAN~\cite{dai2019second} & 15.9 & 34.75 & 0.9300 & 30.59 & 0.8476 & 29.33 & 0.8112 & 28.93 & 0.8671 & 34.30 & 0.9494 \\
					~ & HAN~\cite{niu2020single} & 16.1 & 34.75 & 0.9299 & 30.67 & 0.8483 & 29.32 & 0.8110 & 29.10 & 0.8705 & 34.48 & 0.9500 \\
					~ & NLSA~\cite{mei2021image} & 44.7 & 34.85 & 0.9306 & 30.70 & 0.8485 & 29.34 & 0.8117 & 29.25 & 0.8726 & 34.57 & 0.9508 \\
					~ & IPT$^\dagger$~\cite{chen2021pre} & 115.6 & 34.81 & - & 30.85 & - & 29.38 & - & 29.49 & - & - & - \\
					~ & SwinIR~\cite{liang2021swinir} & 11.9 & 34.91 & 0.9317 & 30.90 & 0.8531 & 29.43 & 0.8140 & 29.65 & 0.8809 & 35.05 & 0.9531 \\
					~ & SwinIR{$^\ddagger$}~\cite{liang2021swinir} & 11.9 & \textcolor{blue}{34.97} & \textcolor{blue}{0.9318} & \textcolor{blue}{30.93} & \textcolor{blue}{0.8534} & \textcolor{blue}{29.46} & \textcolor{blue}{0.8145} & \textcolor{blue}{29.75} & \textcolor{blue}{0.8826} & 35.12 & \textcolor{blue}{0.9537} \\
					~ & EDT-B(Ours) & 11.7 & \textcolor{blue}{34.97} & 0.9316 & 30.89 & 0.8527 & 29.44 & 0.8142 & 29.72 & 0.8814 & \textcolor{blue}{35.13} & 0.9534 \\
					~ & \textbf{EDT-B}$^\dagger$(Ours) & 11.7 & \textcolor{red}{35.13} & \textcolor{red}{0.9328} & \textcolor{red}{31.09} & \textcolor{red}{0.8553} & \textcolor{red}{29.53} & \textcolor{red}{0.8165} & \textcolor{red}{30.07} & \textcolor{red}{0.8863} & \textcolor{red}{35.47} & \textcolor{red}{0.9550} \\
					\hline
					
					\multirow{9}{*}{$\times 4$} & RCAN~\cite{zhang2018image} & 15.6 & 32.63 & 0.9002 & 28.87 & 0.7889 & 27.77 & 0.7436 & 26.82 & 0.8087 & 31.22 & 0.9173 \\
					~ & SAN~\cite{dai2019second} & 15.9 & 32.64 & 0.9003 & 28.92 & 0.7888 & 27.78 & 0.7436 & 26.79 & 0.8068 & 31.18 & 0.9169 \\
					~ & HAN~\cite{niu2020single} & 16.1 & 32.64 & 0.9002 & 28.90 & 0.7890 & 27.80 & 0.7442 & 26.85 & 0.8094 & 31.42 & 0.9177 \\
					~ & NLSA~\cite{mei2021image} & 44.2 & 32.59 & 0.9000 & 28.87 & 0.7891 & 27.78 & 0.7444 & 26.96 & 0.8109 & 31.27 & 0.9184 \\
					~ & IPT$^\dagger$~\cite{chen2021pre} & 115.6 & 32.64 & - & 29.01 & - & 27.82 & - & 27.26 & - & - & - \\
					~ & SwinIR~\cite{liang2021swinir} & 11.9 & 32.74 & 0.9020 & 29.06 & 0.7939 & 27.89 & 0.7479 & 27.37 & 0.8233 & 31.93 & 0.9246 \\
					~ & SwinIR{$^\ddagger$}~\cite{liang2021swinir} & 11.9 & \textcolor{blue}{32.92} & \textcolor{blue}{0.9044} & \textcolor{blue}{29.09} & \textcolor{blue}{0.7950} & \textcolor{blue}{27.92} & \textcolor{blue}{0.7489} & 27.45 & \textcolor{blue}{0.8254} & 32.03 & \textcolor{blue}{0.9260} \\
					~ & EDT-B(Ours) & 11.6 & 32.82 & 0.9031 & \textcolor{blue}{29.09} & 0.7939 & 27.91 & 0.7483 & \textcolor{blue}{27.46} & 0.8246 & \textcolor{blue}{32.05} & 0.9254 \\
					~ & \textbf{EDT-B}$^\dagger$(Ours) & 11.6 & \textcolor{red}{33.06} & \textcolor{red}{0.9055} & \textcolor{red}{29.23} & \textcolor{red}{0.7971} & \textcolor{red}{27.99} & \textcolor{red}{0.7510} & \textcolor{red}{27.75} & \textcolor{red}{0.8317} & \textcolor{red}{32.39} & \textcolor{red}{0.9283} \\
					\hline
				\end{tabular}
			}
		\end{center}
		\vspace{-0.2in}
		\label{tab:classical_sr}
	\end{table}
	
	\begin{table}[t]
		\caption{Quantitative comparison for lightweight SR on PSNR(dB)/SSIM on the Y channel. ``$\ddagger$'' means the $\times 3$ and $\times 4$ models of SwinIR are pre-trained on the $\times 2$ setup and the training patch size is $64 \times 64$ (ours is $48 \times 48$).}
	    \vspace{-0.15in}
		\renewcommand\arraystretch{1.1}
		\begin{center}
			\resizebox{\textwidth}{!}{
				\begin{tabular}{| c | l | c | c | c | c | c | c | c | c | c | c | c |}
					\hline
					\multirow{2}{*}{Scale} & \multirow{2}{*}{Method} & \#Param. & \multicolumn{2}{|c|}{Set5} & \multicolumn{2}{|c|}{Set14} & \multicolumn{2}{|c|}{BSDS100} & \multicolumn{2}{|c|}{Urban100} & \multicolumn{2}{|c|}{Manga109} \\
					\cline{4-13}
					~ & ~ & ($\times 10^3$) & PSNR & SSIM & PSNR & SSIM & PSNR & SSIM & PSNR & SSIM & PSNR & SSIM \\
					\hline
					\multirow{4}{*}{$\times 2$} & LAPAR~\cite{li2020lapar} & 548 & 38.01 & 0.9605 & 33.62 & 0.9183 & 32.19 & 0.8999 & 32.10 & 0.9283 & 38.67 & 0.9772 \\
					~ & LatticeNet~\cite{luo2020latticenet} & 756 & \textcolor{blue}{38.15} & 0.9610 & 33.78 & 0.9193 & 32.25 & 0.9005 & 32.43 & 0.9302 & - & - \\
					~ & SwinIR{$^\ddagger$}~\cite{liang2021swinir} & 878 & 38.14 & \textcolor{blue}{0.9611} & \textcolor{blue}{33.86} & \textcolor{blue}{0.9206} & \textcolor{blue}{32.31} & \textcolor{blue}{0.9012} & \textcolor{blue}{32.76} & \textcolor{blue}{0.9340} & \textcolor{blue}{39.12} & \textcolor{blue}{0.9783} \\
					~ & \textbf{EDT-T}(Ours) & 917 & \textcolor{red}{38.23} & \textcolor{red}{0.9615} & \textcolor{red}{33.99} & \textcolor{red}{0.9209} & \textcolor{red}{32.37} & \textcolor{red}{0.9021} & \textcolor{red}{32.98} & \textcolor{red}{0.9362} & \textcolor{red}{39.45} & \textcolor{red}{0.9789} \\
					\hline
					
					\multirow{4}{*}{$\times 3$} & LAPAR~\cite{li2020lapar} & 594 & 34.36 & 0.9267 & 30.34 & 0.8421 & 29.11 & 0.8054 & 28.15 & 0.8523 & 33.51 & 0.9441 \\
					~ & LatticeNet~\cite{luo2020latticenet} & 765 & 34.53 & 0.9281 & 30.39 & 0.8424 & 29.15 & 0.8059 & 28.33 & 0.8538 & - & - \\
					~ & SwinIR{$^\ddagger$}~\cite{liang2021swinir} & 886 & \textcolor{blue}{34.62} & \textcolor{blue}{0.9289} & \textcolor{blue}{30.54} & \textcolor{blue}{0.8463} & \textcolor{blue}{29.20} & \textcolor{blue}{0.8082} & \textcolor{blue}{28.66} & \textcolor{blue}{0.8624} & \textcolor{blue}{33.98} & \textcolor{blue}{0.9478} \\
					~ & \textbf{EDT-T}(Ours) & 919 & \textcolor{red}{34.73} & \textcolor{red}{0.9299} & \textcolor{red}{30.66} & \textcolor{red}{0.8481} & \textcolor{red}{29.29} & \textcolor{red}{0.8103} & \textcolor{red}{28.89} & \textcolor{red}{0.8674} & \textcolor{red}{34.44} & \textcolor{red}{0.9498} \\
					\hline
					
					\multirow{4}{*}{$\times 4$} & LAPAR~\cite{li2020lapar} & 659 & 32.15 & 0.8944 & 28.61 & 0.7818 & 27.61 & 0.7366 & 26.14 & 0.7871 & 30.42 & 0.9074 \\
					~ & LatticeNet~\cite{luo2020latticenet} & 777 & 32.30 & 0.8962 & 28.68 & 0.7830 & 27.62 & 0.7367 & 26.25 & 0.7873 & - & - \\
					~ & SwinIR{$^\ddagger$}~\cite{liang2021swinir} & 897 & \textcolor{blue}{32.44} & \textcolor{blue}{0.8976} & \textcolor{blue}{28.77} & \textcolor{blue}{0.7858} & \textcolor{blue}{27.69} & \textcolor{blue}{0.7406} & \textcolor{blue}{26.47} & \textcolor{blue}{0.7980} & \textcolor{blue}{30.92} & \textcolor{blue}{0.9151} \\
					~ & \textbf{EDT-T}(Ours) & 922 & \textcolor{red}{32.53} & \textcolor{red}{0.8991} & \textcolor{red}{28.88} & \textcolor{red}{0.7882} & \textcolor{red}{27.76} & \textcolor{red}{0.7433} & \textcolor{red}{26.71} & \textcolor{red}{0.8051} & \textcolor{red}{31.35} & \textcolor{red}{0.9180} \\
					\hline
				\end{tabular}
			}
		\end{center}
		\vspace{-0.05in}
		\label{tab:lightweight_sr}
	\end{table}
	
	\begin{table}[t]
		\caption{Quantitative comparison for color image denoising on PSNR(dB) on RGB channels. ``$\ddagger$'' means the $\sigma=25/50$ models of SwinIR~\cite{liang2021swinir} are pre-trained on the $\sigma=15$ level. ``$\dagger$'' indicates methods with a pre-training. ``$\ast$'' means our model without downsampling and \textit{without pre-training}.}
		\renewcommand\arraystretch{1.1}
		\setlength\tabcolsep{3pt}
		\begin{center}
			\resizebox{\textwidth}{!}{
				\begin{tabular}{| c | c | c | c | c | c | c | c | c | c | c | c | c | c | c |}
					\hline
					\multirow{2}{*}{Dataset} & \multirow{2}{*}{$\sigma$} & BM3D & DnCNN & IRCNN & FFDNet & BRDNet & RDN & IPT$^\dagger$ & DRUNet & SwinIR$^\ddagger$ & \textbf{EDT-B} & \textbf{EDT-B}$^\dagger$ & \textbf{EDT-B}$^\ast$ \\
					~ & ~ & \cite{dabov2007image} & \cite{zhang2017beyond} & \cite{zhang2017learning} & \cite{zhang2018ffdnet} & \cite{tian2020image} & \cite{zhang2020residual} & \cite{chen2021pre} & \cite{zhang2021plug} & \cite{liang2021swinir} & (Ours) & (Ours) & (Ours) \\
					\hline
					\multirow{3}{*}{CBSD68} & 15 & 33.52 & 33.90 & 33.86 & 33.87 & 34.10 & - & - & 34.30 & \textcolor{red}{34.42} & 34.33 & 34.38 & \textcolor{blue}{34.39}\\
					~ & 25 & 30.71 & 31.24 & 31.16 & 31.21 & 31.43 & - & - & 31.69 & \textcolor{red}{31.78} & 31.73 & \textcolor{blue}{31.76} & \textcolor{blue}{31.76}\\
					~ & 50 & 27.38 & 27.95 & 27.86 & 27.96 & 28.16 & 28.31 & 28.39 & 28.51 & \textcolor{blue}{28.56} & 28.55 & \textcolor{red}{28.57} & \textcolor{blue}{28.56} \\
					\hline
					\multirow{3}{*}{Kodak24} & 15 & 34.28 & 34.60 & 34.69 & 34.63 & 34.88 & - & - & 35.31 & \textcolor{blue}{35.34} & 35.25 & 35.31 & \textcolor{red}{35.37} \\
					~ & 25 & 32.15 & 32.14 & 32.18 & 32.13 & 32.41 & - & - & \textcolor{blue}{32.89} & \textcolor{blue}{32.89} & 32.84 & \textcolor{blue}{32.89} & \textcolor{red}{32.94} \\
					~ & 50 & 28.46 & 28.95 & 28.93 & 28.98 & 29.22 & 29.66 & 29.64 & \textcolor{blue}{29.86} & 29.79 & 29.81 & 29.83 & \textcolor{red}{29.87} \\
					\hline
					\multirow{3}{*}{McMaster} & 15 & 34.06 & 33.45 & 34.58 & 34.66 & 35.08 & - & - & 35.40 & \textcolor{red}{35.61} & 35.43 & \textcolor{blue}{35.51} & \textcolor{red}{35.61} \\
					~ & 25 & 31.66 & 31.52 & 32.18 & 32.35 & 32.75 & - & - & 33.14 & 33.20 & 33.20 & \textcolor{blue}{33.26} & \textcolor{red}{33.34} \\
					~ & 50 & 28.51 & 28.62 & 28.91 & 29.18 & 29.52 & - & 29.98 & 30.08 & \textcolor{blue}{30.22} & 30.21 & \textcolor{red}{30.25} & \textcolor{red}{30.25} \\
					\hline
					\multirow{3}{*}{Urban100} & 15 & 33.93 & 32.98 & 33.78 & 33.83 & 34.42 & - & - & 34.81 & \textcolor{blue}{35.13} & 34.93 & 35.04 & \textcolor{red}{35.22} \\
					~ & 25 & 31.36 & 30.81 & 31.20 & 31.40 & 31.99 & - & - & 32.60 & \textcolor{blue}{32.90} & 32.78 & 32.86 & \textcolor{red}{33.07} \\
					~ & 50 & 27.93 & 27.59 & 27.70 & 28.05 & 28.56 & 29.38 & 29.71 & 29.61 & 29.82 & 29.93 & \textcolor{blue}{29.98} & \textcolor{red}{30.16} \\
					\hline
				\end{tabular}
			}
		\end{center}
		\vspace{-0.25in}
		\label{tab:denoise}
	\end{table}
	
	\subsection{Denoising Results}
	\label{sec:denoising}
	
	In Table~\ref{tab:denoise}, apart from a range of denoising methods, we present our three models: (1) EDT-B without pre-training; (2) EDT-B with pre-training; (3) EDT-B without downsampling and pre-training.
	
	It is worthwhile to note that, unlike SR models that benefit a lot from pre-training, denoising models only achieve 0.02-0.11dB gains. One possible reason is that we use a large training dataset in denoising tasks, which already provides sufficient data to make the capacity of our models into full play. On the other hand, pre-training hardly affects the internal feature representation of models, discussed in Sec.~\ref{sec:finetune}. Therefore, we suggest that the Gaussian denoising task may not need a large amount of training data.
	
	Besides, we find our encoder-decoder-based framework is well performed on high noise levels (e.g., $\sigma=50$), while yielding slightly inferior performance on low noise levels (e.g., $\sigma=15$). This could be caused by the downsampling operation in EDT. To verify this assumption, we train another EDT-B model without downsampling. As shown in Table~\ref{tab:denoise}, it does obtain better performance on the low level noises. Nonetheless, we suggest that the proposed EDT model is still a good choice for denoising tasks since \textit{it strikes a sweet point between performance and computational complexity}. For example, the FLOPs of EDT-B (38G) is only \textit{8.4\%} of SwinIR (451G).

	\subsection{Deraining Results}
	
	We also evaluate the performance of the proposed EDT on Rain100L~\cite{yang2019joint} and Rain100H~\cite{yang2019joint} two benchmark datasets, accounting for light and heavy rain streaks. As illustrated in Table~\ref{tab:derain}, though the model size of our EDT-B (11.5M) for deraining is far smaller than IPT (116M), it still outperforms IPT by \textit{0.52}dB on the light rain setting. Meanwhile, our model reaches significantly superior results by \textit{2.66}dB gain on the heavy rain setting, compared to the second-best RCDNet~\cite{wang2020model}, supporting that EDT performs well for restoration tasks with heavy degradation.

	\begin{table}[t]
		\caption{Quantitative comparison for image deraining on PSNR(dB)/SSIM on the Y channel. ``$\dagger$'' indicates methods with a pre-training.}
		\vspace{-0.1in}
		\renewcommand\arraystretch{1.25}
		\begin{center}
			\resizebox{\linewidth}{!}{
				\begin{tabular}{| c | c | c | c | c | c | c |}
					\hline
					Method & DSC~\cite{luo2015removing} & GMM~\cite{li2016rain} & JCAS~\cite{gu2017joint} & Clear~\cite{fu2017clearing} & DDN~\cite{fu2017removing} & RESCAN~\cite{li2018recurrent} \\
					\hline
					RAIN100L &  27.34/0.8494 & 29.05/0.8717 & 28.54/0.8524 & 30.24/0.9344 & 32.38/0.9258 & 38.52/0.9812 \\
					\hline
					RAIN100H & 13.77/0.3199 & 15.23/0.4498 & 14.62/0.4510 & 15.33/0.7421 & 22.85/0.7250 & 29.62/0.8720 \\
					\hline
					\hline
					PReNet~\cite{ren2019progressive} & SPANet~\cite{wang2019spatial} & JORDER~\cite{yang2019joint} & SSIR~\cite{wei2019semi} & RCDNet~\cite{wang2020model} & IPT$^\dagger$~\cite{chen2021pre} & \textbf{EDT-B}$^\dagger$(Ours) \\
					\hline
					37.45/0.9790 & 35.33/0.9694 & 38.59/0.9834 & 32.37/0.9258 & 40.00/0.9860 & \textcolor{blue}{41.62}/\textcolor{blue}{0.9880} & \textcolor{red}{42.14}/\textcolor{red}{0.9903} \\
					\hline
					30.11/0.9053 & 25.11/0.8332 & 30.50/0.8967 & 22.47/0.7164 & \textcolor{blue}{31.28}/\textcolor{blue}{0.9093} & -/- & \textcolor{red}{33.94}/\textcolor{red}{0.9398} \\
					
					\hline
				\end{tabular}
			}
		\end{center}
		\vspace{-0.15in}
		\label{tab:derain}
		
	\end{table}
	
	
	\section{Limitations}
	In this paper, we mainly investigate the effect of pre-training and how to conduct effective pre-training on synthesized data. Future research should be undertaken to explore the real-world setting and more tasks, further extended to video processing. Also, despite the high efficiency of our encoder-decoder design, a limitation remains that we adopt a fixed downsampling strategy. As shown in Sec.~\ref{sec:denoising}, it may be a better choice to conduct adaptive downsampling based on degradation degrees of low-quality images.
	
	
	\section{Conclusion}
	Based on the proposed encoder-decoder-based framework that shows high efficiency and strong performance, we perform an in-depth analysis of transformer-based image pre-training in low-level vision. We find pre-training plays the central role of developing stronger intermediate representations by incorporating more local information. Also, we find the effect of pre-training is task-specific, leading to significant improvements on SR and deraining while limited gains on denoising. Then, we suggest multi-related-task pre-training exhibits great potential in digging image priors, far more efficient than using larger pre-training datasets. Finally, we show how data scale and model size affect the performance of pre-training and present comparisons between transformers and ConvNets.
	
	
	
	\clearpage
	%
	%
	\bibliographystyle{splncs04}
	\bibliography{egbib_pretrain}

\begin{thebibliography}{10}
\providecommand{\url}[1]{\texttt{#1}}
\providecommand{\urlprefix}{URL }
\providecommand{\doi}[1]{https://doi.org/#1}

\bibitem{agustsson2017ntire}
Agustsson, E., Timofte, R.: Ntire 2017 challenge on single image
  super-resolution: Dataset and study. In: CVPRW. pp. 126--135 (2017)

\bibitem{arbelaez2010contour}
Arbelaez, P., Maire, M., Fowlkes, C., Malik, J.: Contour detection and
  hierarchical image segmentation. PAMI  \textbf{33}(5),  898--916 (2010)

\bibitem{bao2021beit}
Bao, H., Dong, L., Wei, F.: Beit: Bert pre-training of image transformers.
  arXiv preprint arXiv:2106.08254  (2021)

\bibitem{bevilacqua2012low}
Bevilacqua, M., Roumy, A., Guillemot, C., line Alberi~Morel, M.: Low-complexity
  single-image super-resolution based on nonnegative neighbor embedding. In:
  Proceedings of the British Machine Vision Conference. pp. 135.1--135.10. BMVA
  Press (2012)

\bibitem{brown2020language}
Brown, T.B., Mann, B., Ryder, N., Subbiah, M., Kaplan, J., Dhariwal, P.,
  Neelakantan, A., Shyam, P., Sastry, G., Askell, A., et~al.: Language models
  are few-shot learners. arXiv preprint arXiv:2005.14165  (2020)

\bibitem{carion2020end}
Carion, N., Massa, F., Synnaeve, G., Usunier, N., Kirillov, A., Zagoruyko, S.:
  End-to-end object detection with transformers. In: ECCV. pp. 213--229.
  Springer (2020)

\bibitem{chen2021pre}
Chen, H., Wang, Y., Guo, T., Xu, C., Deng, Y., Liu, Z., Ma, S., Xu, C., Xu, C.,
  Gao, W.: Pre-trained image processing transformer. In: CVPR. pp. 12299--12310
  (2021)

\bibitem{chen2017deeplab}
Chen, L.C., Papandreou, G., Kokkinos, I., Murphy, K., Yuille, A.L.: Deeplab:
  Semantic image segmentation with deep convolutional nets, atrous convolution,
  and fully connected crfs. PAMI  \textbf{40}(4),  834--848 (2017)

\bibitem{cortes2012algorithms}
Cortes, C., Mohri, M., Rostamizadeh, A.: Algorithms for learning kernels based
  on centered alignment. The Journal of Machine Learning Research
  \textbf{13}(1),  795--828 (2012)

\bibitem{dabov2007image}
Dabov, K., Foi, A., Katkovnik, V., Egiazarian, K.: Image denoising by sparse
  3-d transform-domain collaborative filtering. TIP  \textbf{16}(8),
  2080--2095 (2007)

\bibitem{dai2019second}
Dai, T., Cai, J., Zhang, Y., Xia, S.T., Zhang, L.: Second-order attention
  network for single image super-resolution. In: CVPR. pp. 11065--11074 (2019)

\bibitem{deng2009imagenet}
Deng, J., Dong, W., Socher, R., Li, L.J., Li, K., Fei-Fei, L.: Imagenet: A
  large-scale hierarchical image database. In: CVPR. pp. 248--255. Ieee (2009)

\bibitem{devlin2018bert}
Devlin, J., Chang, M.W., Lee, K., Toutanova, K.: Bert: Pre-training of deep
  bidirectional transformers for language understanding. arXiv preprint
  arXiv:1810.04805  (2018)

\bibitem{donahue2014decaf}
Donahue, J., Jia, Y., Vinyals, O., Hoffman, J., Zhang, N., Tzeng, E., Darrell,
  T.: Decaf: A deep convolutional activation feature for generic visual
  recognition. In: ICML. pp. 647--655. PMLR (2014)

\bibitem{dong2021cswin}
Dong, X., Bao, J., Chen, D., Zhang, W., Yu, N., Yuan, L., Chen, D., Guo, B.:
  Cswin transformer: A general vision transformer backbone with cross-shaped
  windows. arXiv preprint arXiv:2107.00652  (2021)

\bibitem{dosovitskiy2020image}
Dosovitskiy, A., Beyer, L., Kolesnikov, A., Weissenborn, D., Zhai, X.,
  Unterthiner, T., Dehghani, M., Minderer, M., Heigold, G., Gelly, S., et~al.:
  An image is worth 16x16 words: Transformers for image recognition at scale.
  In: ICLR (2020)

\bibitem{richdodak}
Franzen, R.: Kodak lossless true color image suite.,
  \url{http://r0k.us/graphics/kodak/}

\bibitem{fu2017clearing}
Fu, X., Huang, J., Ding, X., Liao, Y., Paisley, J.: Clearing the skies: A deep
  network architecture for single-image rain removal. TIP  \textbf{26}(6),
  2944--2956 (2017)

\bibitem{fu2017removing}
Fu, X., Huang, J., Zeng, D., Huang, Y., Ding, X., Paisley, J.: Removing rain
  from single images via a deep detail network. In: CVPR. pp. 3855--3863 (2017)

\bibitem{girshick2015fast}
Girshick, R.: Fast r-cnn. In: ICCV. pp. 1440--1448 (2015)

\bibitem{girshick2014rich}
Girshick, R., Donahue, J., Darrell, T., Malik, J.: Rich feature hierarchies for
  accurate object detection and semantic segmentation. In: CVPR. pp. 580--587
  (2014)

\bibitem{gretton2007kernel}
Gretton, A., Fukumizu, K., Teo, C.H., Song, L., Sch{\"o}lkopf, B., Smola, A.J.,
  et~al.: A kernel statistical test of independence. In: Nips. vol.~20, pp.
  585--592. Citeseer (2007)

\bibitem{gu2021interpreting}
Gu, J., Dong, C.: Interpreting super-resolution networks with local attribution
  maps. In: Proceedings of the IEEE/CVF Conference on Computer Vision and
  Pattern Recognition. pp. 9199--9208 (2021)

\bibitem{gu2017joint}
Gu, S., Meng, D., Zuo, W., Zhang, L.: Joint convolutional analysis and
  synthesis sparse representation for single image layer separation. In: ICCV.
  pp. 1708--1716 (2017)

\bibitem{he2021masked}
He, K., Chen, X., Xie, S., Li, Y., Doll{\'a}r, P., Girshick, R.: Masked
  autoencoders are scalable vision learners. arXiv preprint arXiv:2111.06377
  (2021)

\bibitem{huang2015single}
Huang, J.B., Singh, A., Ahuja, N.: Single image super-resolution from
  transformed self-exemplars. In: CVPR. pp. 5197--5206 (2015)

\bibitem{kolesnikov2020big}
Kolesnikov, A., Beyer, L., Zhai, X., Puigcerver, J., Yung, J., Gelly, S.,
  Houlsby, N.: Big transfer (bit): General visual representation learning. In:
  ECCV. pp. 491--507. Springer (2020)

\bibitem{kornblith2019better}
Kornblith, S., Shlens, J., Le, Q.V.: Do better imagenet models transfer better?
  In: CVPR. pp. 2661--2671 (2019)

\bibitem{lecun1989backpropagation}
LeCun, Y., Boser, B., Denker, J.S., Henderson, D., Howard, R.E., Hubbard, W.,
  Jackel, L.D.: Backpropagation applied to handwritten zip code recognition.
  Neural computation  \textbf{1}(4),  541--551 (1989)

\bibitem{li2020lapar}
Li, W., Zhou, K., Qi, L., Jiang, N., Lu, J., Jia, J.: Lapar: Linearly-assembled
  pixel-adaptive regression network for single image super-resolution and
  beyond. NeurIPS  \textbf{33} (2020)

\bibitem{li2018recurrent}
Li, X., Wu, J., Lin, Z., Liu, H., Zha, H.: Recurrent squeeze-and-excitation
  context aggregation net for single image deraining. In: ECCV. pp. 254--269
  (2018)

\bibitem{li2016rain}
Li, Y., Tan, R.T., Guo, X., Lu, J., Brown, M.S.: Rain streak removal using
  layer priors. In: CVPR. pp. 2736--2744 (2016)

\bibitem{liang2021swinir}
Liang, J., Cao, J., Sun, G., Zhang, K., Van~Gool, L., Timofte, R.: Swinir:
  Image restoration using swin transformer. In: ICCVW. pp. 1833--1844 (2021)

\bibitem{liu2021Swin}
Liu, Z., Lin, Y., Cao, Y., Hu, H., Wei, Y., Zhang, Z., Lin, S., Guo, B.: Swin
  transformer: Hierarchical vision transformer using shifted windows. ICCV
  (2021)

\bibitem{long2015fully}
Long, J., Shelhamer, E., Darrell, T.: Fully convolutional networks for semantic
  segmentation. In: CVPR. pp. 3431--3440 (2015)

\bibitem{luo2020latticenet}
Luo, X., Xie, Y., Zhang, Y., Qu, Y., Li, C., Fu, Y.: Latticenet: Towards
  lightweight image super-resolution with lattice block. In: ECCV. pp.
  272--289. Springer (2020)

\bibitem{luo2015removing}
Luo, Y., Xu, Y., Ji, H.: Removing rain from a single image via discriminative
  sparse coding. In: ICCV. pp. 3397--3405 (2015)

\bibitem{ma2016waterloo}
Ma, K., Duanmu, Z., Wu, Q., Wang, Z., Yong, H., Li, H., Zhang, L.: Waterloo
  exploration database: New challenges for image quality assessment models. TIP
   \textbf{26}(2),  1004--1016 (2016)

\bibitem{mahajan2018exploring}
Mahajan, D., Girshick, R., Ramanathan, V., He, K., Paluri, M., Li, Y.,
  Bharambe, A., Van Der~Maaten, L.: Exploring the limits of weakly supervised
  pretraining. In: ECCV. pp. 181--196 (2018)

\bibitem{martin2001database}
Martin, D., Fowlkes, C., Tal, D., Malik, J.: A database of human segmented
  natural images and its application to evaluating segmentation algorithms and
  measuring ecological statistics. In: ICCV. vol.~2, pp. 416--423. IEEE (2001)

\bibitem{matsui2017sketch}
Matsui, Y., Ito, K., Aramaki, Y., Fujimoto, A., Ogawa, T., Yamasaki, T.,
  Aizawa, K.: Sketch-based manga retrieval using manga109 dataset. Multimedia
  Tools and Applications  \textbf{76}(20),  21811--21838 (2017)

\bibitem{mei2021image}
Mei, Y., Fan, Y., Zhou, Y.: Image super-resolution with non-local sparse
  attention. In: CVPR. pp. 3517--3526 (2021)

\bibitem{nguyen2020wide}
Nguyen, T., Raghu, M., Kornblith, S.: Do wide and deep networks learn the same
  things? uncovering how neural network representations vary with width and
  depth. arXiv preprint arXiv:2010.15327  (2020)

\bibitem{niu2020single}
Niu, B., Wen, W., Ren, W., Zhang, X., Yang, L., Wang, S., Zhang, K., Cao, X.,
  Shen, H.: Single image super-resolution via a holistic attention network. In:
  ECCV. pp. 191--207. Springer (2020)

\bibitem{radford2018improving}
Radford, A., Narasimhan, K., Salimans, T., Sutskever, I.: Improving language
  understanding by generative pre-training  (2018)

\bibitem{radford2019language}
Radford, A., Wu, J., Child, R., Luan, D., Amodei, D., Sutskever, I., et~al.:
  Language models are unsupervised multitask learners. OpenAI blog
  \textbf{1}(8), ~9 (2019)

\bibitem{raffel2020exploring}
Raffel, C., Shazeer, N., Roberts, A., Lee, K., Narang, S., Matena, M., Zhou,
  Y., Li, W., Liu, P.J.: Exploring the limits of transfer learning with a
  unified text-to-text transformer. Journal of Machine Learning Research
  \textbf{21},  1--67 (2020)

\bibitem{raghu2021vision}
Raghu, M., Unterthiner, T., Kornblith, S., Zhang, C., Dosovitskiy, A.: Do
  vision transformers see like convolutional neural networks? arXiv preprint
  arXiv:2108.08810  (2021)

\bibitem{ren2019progressive}
Ren, D., Zuo, W., Hu, Q., Zhu, P., Meng, D.: Progressive image deraining
  networks: A better and simpler baseline. In: CVPR. pp. 3937--3946 (2019)

\bibitem{sharif2014cnn}
Sharif~Razavian, A., Azizpour, H., Sullivan, J., Carlsson, S.: Cnn features
  off-the-shelf: an astounding baseline for recognition. In: CVPRW. pp.
  806--813 (2014)

\bibitem{strudel2021segmenter}
Strudel, R., Garcia, R., Laptev, I., Schmid, C.: Segmenter: Transformer for
  semantic segmentation. arXiv preprint arXiv:2105.05633  (2021)

\bibitem{sun2017revisiting}
Sun, C., Shrivastava, A., Singh, S., Gupta, A.: Revisiting unreasonable
  effectiveness of data in deep learning era. In: ICCV. pp. 843--852 (2017)

\bibitem{tian2020image}
Tian, C., Xu, Y., Zuo, W.: Image denoising using deep cnn with batch
  renormalization. Neural Networks  \textbf{121},  461--473 (2020)

\bibitem{timofte2017ntire}
Timofte, R., Agustsson, E., Van~Gool, L., Yang, M.H., Zhang, L.: Ntire 2017
  challenge on single image super-resolution: Methods and results. In: CVPRW.
  pp. 114--125 (2017)

\bibitem{vaswani2021scaling}
Vaswani, A., Ramachandran, P., Srinivas, A., Parmar, N., Hechtman, B., Shlens,
  J.: Scaling local self-attention for parameter efficient visual backbones.
  In: CVPR. pp. 12894--12904 (2021)

\bibitem{vaswani2017attention}
Vaswani, A., Shazeer, N., Parmar, N., Uszkoreit, J., Jones, L., Gomez, A.N.,
  Kaiser, {\L}., Polosukhin, I.: Attention is all you need. In: NIPS. pp.
  5998--6008 (2017)

\bibitem{wang2020model}
Wang, H., Xie, Q., Zhao, Q., Meng, D.: A model-driven deep neural network for
  single image rain removal. In: CVPR. pp. 3103--3112 (2020)

\bibitem{wang2019spatial}
Wang, T., Yang, X., Xu, K., Chen, S., Zhang, Q., Lau, R.W.: Spatial attentive
  single-image deraining with a high quality real rain dataset. In: CVPR. pp.
  12270--12279 (2019)

\bibitem{wang2021pyramid}
Wang, W., Xie, E., Li, X., Fan, D.P., Song, K., Liang, D., Lu, T., Luo, P.,
  Shao, L.: Pyramid vision transformer: A versatile backbone for dense
  prediction without convolutions. arXiv preprint arXiv:2102.12122  (2021)

\bibitem{wang2018esrgan}
Wang, X., Yu, K., Wu, S., Gu, J., Liu, Y., Dong, C., Qiao, Y., Change~Loy, C.:
  Esrgan: Enhanced super-resolution generative adversarial networks. In: ECCVW.
  pp.~0--0 (2018)

\bibitem{wang2021uformer}
Wang, Z., Cun, X., Bao, J., Liu, J.: Uformer: A general u-shaped transformer
  for image restoration. arXiv preprint arXiv:2106.03106  (2021)

\bibitem{wei2019semi}
Wei, W., Meng, D., Zhao, Q., Xu, Z., Wu, Y.: Semi-supervised transfer learning
  for image rain removal. In: CVPR. pp. 3877--3886 (2019)

\bibitem{xiao2021early}
Xiao, T., Singh, M., Mintun, E., Darrell, T., Doll{\'a}r, P., Girshick, R.:
  Early convolutions help transformers see better. arXiv preprint
  arXiv:2106.14881  (2021)

\bibitem{xie2021simmim}
Xie, Z., Zhang, Z., Cao, Y., Lin, Y., Bao, J., Yao, Z., Dai, Q., Hu, H.:
  Simmim: A simple framework for masked image modeling. arXiv preprint
  arXiv:2111.09886  (2021)

\bibitem{yang2019joint}
Yang, W., Tan, R.T., Feng, J., Guo, Z., Yan, S., Liu, J.: Joint rain detection
  and removal from a single image with contextualized deep networks. PAMI
  \textbf{42}(6),  1377--1393 (2019)

\bibitem{zeyde2010single}
Zeyde, R., Elad, M., Protter, M.: On single image scale-up using
  sparse-representations. In: International conference on curves and surfaces.
  pp. 711--730. Springer (2010)

\bibitem{zhang2021plug}
Zhang, K., Li, Y., Zuo, W., Zhang, L., Van~Gool, L., Timofte, R.: Plug-and-play
  image restoration with deep denoiser prior. PAMI  (2021)

\bibitem{zhang2017beyond}
Zhang, K., Zuo, W., Chen, Y., Meng, D., Zhang, L.: Beyond a gaussian denoiser:
  Residual learning of deep cnn for image denoising. TIP  \textbf{26}(7),
  3142--3155 (2017)

\bibitem{zhang2017learning}
Zhang, K., Zuo, W., Gu, S., Zhang, L.: Learning deep cnn denoiser prior for
  image restoration. In: CVPR. pp. 3929--3938 (2017)

\bibitem{zhang2018ffdnet}
Zhang, K., Zuo, W., Zhang, L.: Ffdnet: Toward a fast and flexible solution for
  cnn-based image denoising. TIP  \textbf{27}(9),  4608--4622 (2018)

\bibitem{zhang2011color}
Zhang, L., Wu, X., Buades, A., Li, X.: Color demosaicking by local directional
  interpolation and nonlocal adaptive thresholding. Journal of Electronic
  imaging  \textbf{20}(2),  023016 (2011)

\bibitem{zhang2021multi}
Zhang, P., Dai, X., Yang, J., Xiao, B., Yuan, L., Zhang, L., Gao, J.:
  Multi-scale vision longformer: A new vision transformer for high-resolution
  image encoding. arXiv preprint arXiv:2103.15358  (2021)

\bibitem{zhang2018image}
Zhang, Y., Li, K., Li, K., Wang, L., Zhong, B., Fu, Y.: Image super-resolution
  using very deep residual channel attention networks. In: ECCV. pp. 286--301
  (2018)

\bibitem{zhang2020residual}
Zhang, Y., Tian, Y., Kong, Y., Zhong, B., Fu, Y.: Residual dense network for
  image restoration. PAMI  \textbf{43}(7),  2480--2495 (2020)

\end{thebibliography}
	
	\clearpage
	
	\renewcommand\thesection{\Alph{section}}
	\renewcommand\thesubsection{\thesection.\arabic{subsection}}
	\renewcommand\thefigure{\Alph{section}.\arabic{figure}}
	\renewcommand\thetable{\Alph{section}.\arabic{table}} 
	
	\setcounter{section}{0}
	\setcounter{figure}{0}
	\setcounter{table}{0}
	
	\begin{center}
		\noindent{\Large{\textbf{On Efficient Transformer-Based Image Pre-training for Low-Level Vision \\ 
	    \vspace{0.1in}
		(Supplementary Material)}}}
	\end{center}
	\vspace{0.2in}
	
	\section{Network Architecture}
	
\begin{figure}[h]
		\begin{center}
			\includegraphics[width=0.74\linewidth]{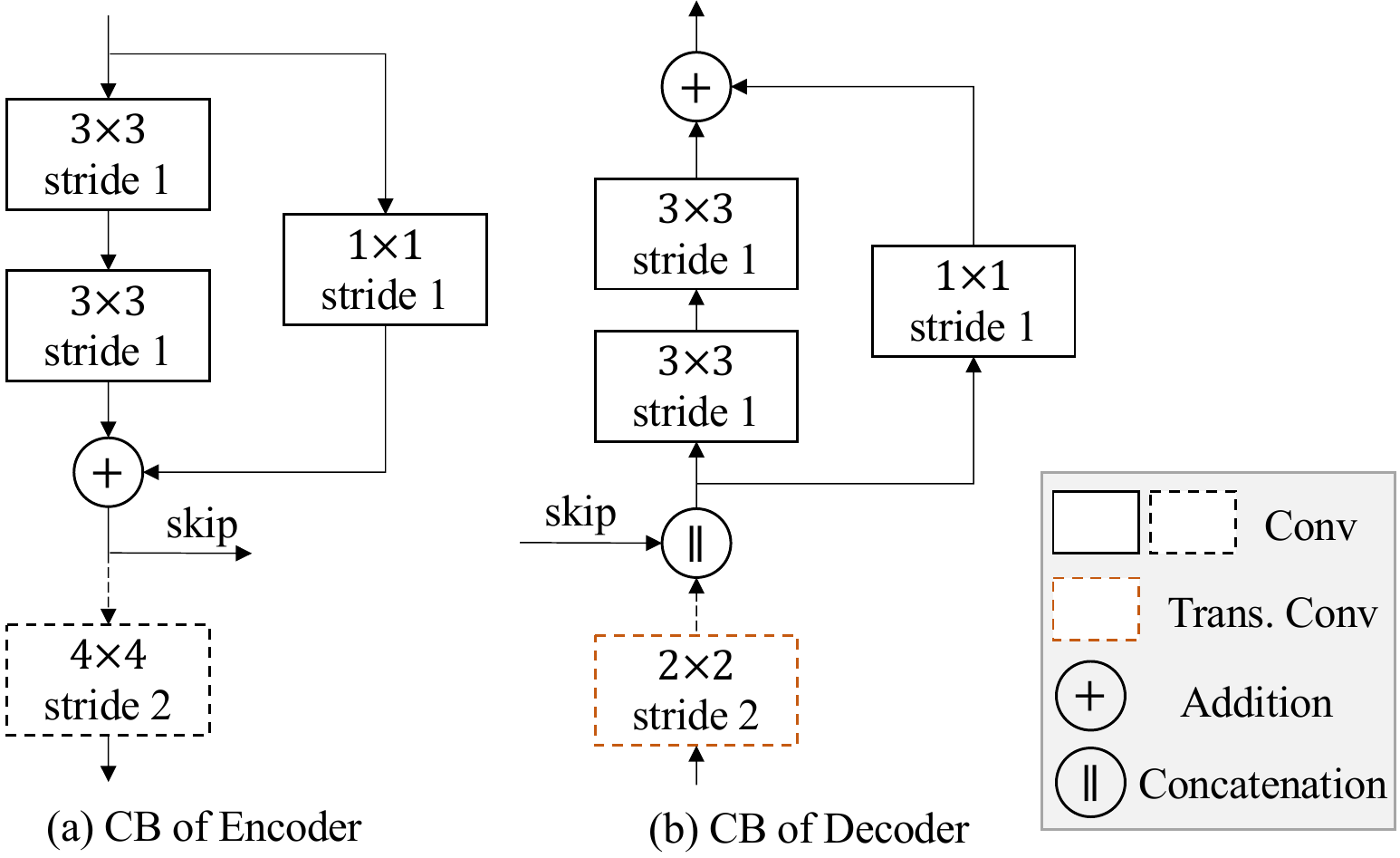}
		\end{center}
		\caption{Structures of the convolution blocks (CB) in the encoder and decoder. The dotted boxes and lines represent additional downsampling and upsampling operations when processing high-resolution inputs.}
		\label{fig:cb}
		
	\end{figure}

\begin{figure}[t]
		\begin{center}
			\includegraphics[width=0.7\linewidth]{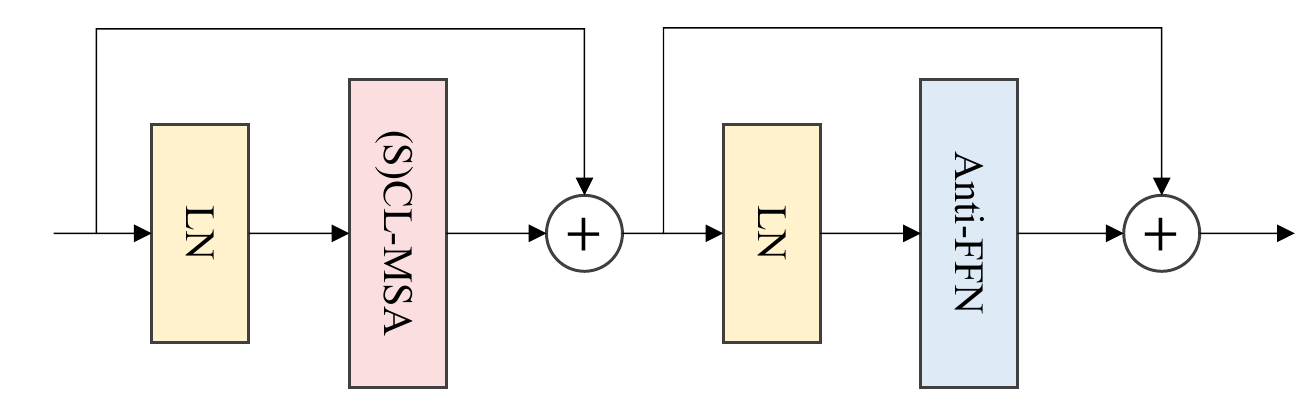}
		\end{center}
		\caption{Structure of the transformer block, including layer normalizations (LN), a (shifted) crossed local multi-head attention module ((S)CL-MSA) and an anti-blocking feed-forward network (Anti-FFN).}
		\label{fig:tb}
	\end{figure}

	The proposed encoder-decoder-based transformer (EDT) in Fig.~{1} is composed of a convolution encoder and decoder as well as a transformer body. It processes low-resolution (\eg, in super-resolution) and high-resolution (\eg, in denoising) inputs using different encoders and decoders, where the high-resolution path involves additional downsampling and upsampling operations, as illustrated in Sec.~{2.1} and Fig.~\ref{fig:cb}. This design enables the transformer body to model long-range relations at a low resolution, thus being computationally efficient. The body consists of multiple stages of transformer blocks, using a global connection in each stage. We show the structure of a transformer block in Fig.~\ref{fig:tb}, where the (shifted) crossed local attention and anti-blocking FFN are detailed in the following. We design four variants of EDT and the corresponding configurations are detailed in Table~{1}. 
	
	\vspace{0.05in}
	\noindent\textbf{Shifted Crossed Local Attention}.
	To reduce computational complexity, several works~\cite{liu2021Swin,vaswani2021scaling} attempt to leverage shifted or halo windows to perform local self-attention. Whereas, the slow growth of effective receptive fields hampers the representational capability. Later on,~\cite{dong2021cswin} proposes to use globally horizontal and vertical stripe attention to achieve global receptive field. However, the computation is a heavy burden when processing high-resolution images. Drawing insights from these works~\cite{liu2021Swin,dong2021cswin}, we design a local self-attention mechanism with shifted crossed windows that strikes a balance of receptive field growth and computational cost, showing competitive performance.
	
	\begin{figure}[t]
		\begin{center}
			\includegraphics[width=0.83\linewidth]{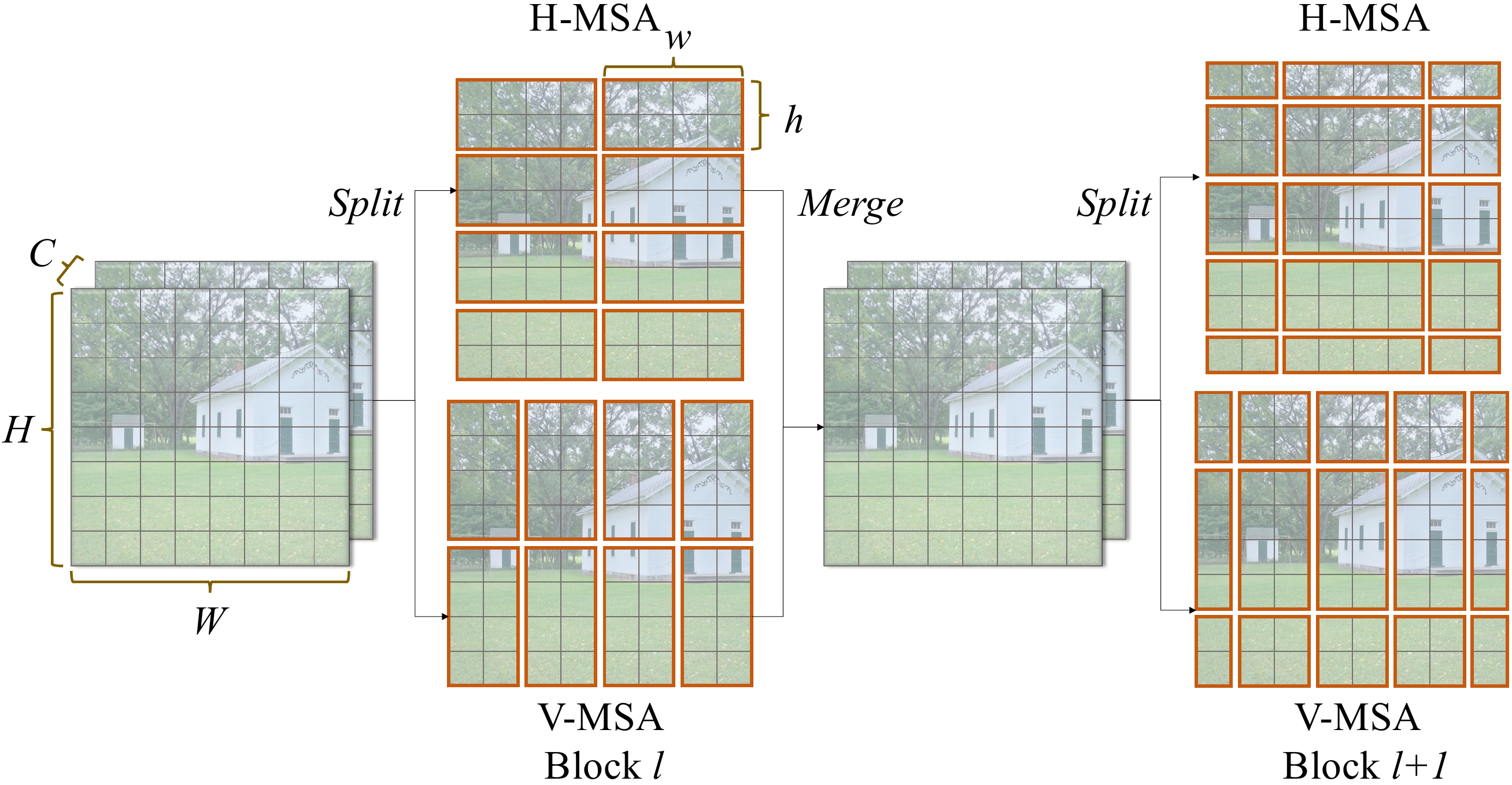}
		\end{center}
		\caption{Shifted Crossed Local Attention. The horizontal and vertical window sizes are set to $(h, w)$ and $(w, h)$. We set $(h, w)$ to (6, 24) in our implementation.}
		\vspace{-0.2in}
		\label{fig:att}
	\end{figure}
	
	As shown in Fig.~\ref{fig:att}, by evenly splitting a given feature map $ \mathbf{X} \in {\mathbb R}^{(H \times W) \times C}$ into two parts in the channel dimension, each half performs the multi-head attention (MSA) in an either horizontal or vertical local window, where the window size is $(h, w)$ or $(w, h)$. Generally speaking, there exists $w \, \textgreater \, h$. The (shifted) crossed local attention (S)CL-MSA is formally defined as:
	\begin{align}
	& \mathbf{X} = [ \mathbf{X}_{1}, \mathbf{X}_{2} ], \ {\rm where} \ \mathbf{X}_{1}, \mathbf{X}_{2} \in {\mathbb R}^{(H \times W) \times \sfrac{C}{2}} \,,
	\end{align}
	\begin{align}
	& \mathbf{X}_{1}^{'} = {\rm H\text{-}MSA} (\mathbf{X}_{1}) \,, \\
	& \mathbf{X}_{2}^{'} = {\rm V\text{-}MSA} (\mathbf{X}_{2}) \,, \\
	& {\rm (S)CL\text{-}MSA} (\mathbf{X}) = {\rm Proj}([ \mathbf{X}_{1}^{'}, \mathbf{X}_{2}^{'} ]) \,,
	\end{align}
	where the projection layer fuses the attention results. Then, in the next transformer block, we shift the horizontal and vertical windows by $(\lfloor \frac{h}{2} \rfloor, \lfloor \frac{w}{2} \rfloor)$ and $(\lfloor \frac{w}{2} \rfloor, \lfloor \frac{h}{2} \rfloor)$ pixels, respectively. The crossed windows with shifts dramatically increase the effective receptive field. We show the fast growth of the receptive field is beneficial in achieving higher restoration quality in Sec.~\ref{sec:abl}. The computational complexity of our attention module for an $H \times W$ image is:
	\begin{align}
	\Omega \left({\rm (S)CW\text{-}MSA} \right) = 4HWC^{2}+ 2hwHWC \,.
	\end{align}

	\noindent\textbf{Anti-Blocking FFN}. 
	To eliminate the possible blocking effect caused by window partition, we design an anti-blocking feed-forward network (Anti-FFN), which is formulated as:
	\begin{align}
	&\mathbf{X}^{'} = {\rm Act}( {\rm Linear}(\mathbf{X}) ) \,, \\
	&\mathbf{X} = {\rm Linear}({\rm Act}( {\rm Anti\text{-}Block}(\mathbf{X}^{'})) ) \,,
	\end{align}
	where the anti-blocking operation is implemented with a depth-wise convolution with a large kernel size ($5 \times 5$). Note that \cite{wang2021uformer} adopts a similar strategy to leverage local context. We further explore the role of this operation in low-level tasks, which is detailed in Sec.~\ref{sec:abl}.
	\vspace{-0.05in}
	
	\section{Study on the Network Architecture}
	\label{sec:abl}
	
\begin{figure}[t]
		\centering
		\begin{minipage}[c]{1.0\linewidth}
			\centering
			\caption{Study on window sizes of the attention block on PSNR(dB) in $\times 2$ SR. Best results are in \textbf{bold}.}
			\vspace{0.15in}
			\renewcommand\arraystretch{1.1}
			\begin{tabular}{| c | c | c | c | c | c |}
				\hline
				Size & Set5 & Set14 & BSDS100 & Urban100 & Manga109 \\
				\hline
				(8, 8) & 38.37 & 34.51 & 32.48 & 33.57 & 39.82 \\
				(4, 16) & 38.39 & 34.48 & 32.49 & 33.66 & 39.83 \\
				(8, 16) & 38.42 & 34.53 & 32.51 & 33.78 & 39.92 \\
				(12, 12) & \textbf{38.45} & 34.56 & 32.50 & 33.73 & 39.89 \\
				(6, 24) & \textbf{38.45} & \textbf{34.57} & \textbf{32.52} & \textbf{33.80} & \textbf{39.93} \\				
				\hline
			\end{tabular}
			\label{tab:window}
		\end{minipage}
		
		\vspace{0.2in}
		\begin{minipage}[c]{1.0\linewidth}
			\centering
			\includegraphics[width=0.6\textwidth]{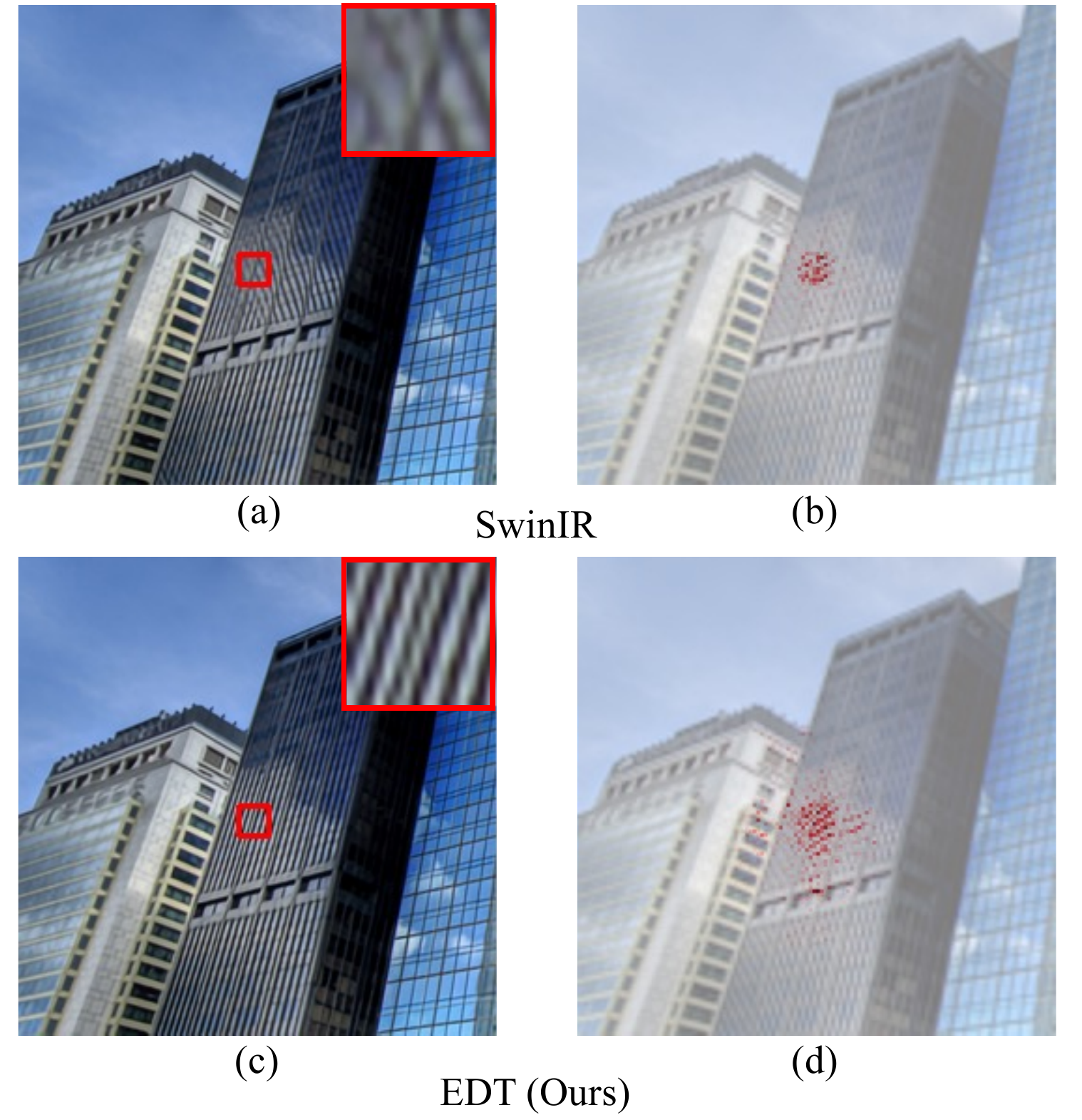}
			\captionof{figure}{LAM~\cite{gu2021interpreting} comparison between SwinIR~\cite{liang2021swinir} and EDT in $\times 2$ SR. The first column shows input along with the super-resolved result, and the second column shows the receptive field.}
			\label{fig:Rf}
		\end{minipage}
		\vspace{-0.2in}
	\end{figure}

	\noindent{\textbf{Window Size.}} To evaluate the effectiveness of the proposed shifted crossed local attention, we conduct experiments to analyze the effects of different window sizes on final performance. As shown in Table~\ref{tab:window}, when simply increasing the square window size from 8 to 12, the results of five SR benchmarks are largely improved, verifying the importance of a large receptive field under the current architecture. The evidence also comes from the comparison between (4, 16) and (8, 16), where we find a larger shot side achieves superior performance. Besides, for windows with the same area, (6, 24) is better than (12, 12), indicating our shifted crossed window attention is an effective way of quickly enlarging the receptive field to improve performance. This claim is also supported in Table~{5}, where our lightweight SR models (EDT-T) yield significant improvements on high-resolution benchmark datasets Urban100~\cite{huang2015single} and Manga109~\cite{matsui2017sketch}. Further, we use LAM~\cite{gu2021interpreting}, which represents the range of information utilization, to visualize receptive fields. Fig.~\ref{fig:Rf} shows our model can take advantage of a wider range of information than SwinIR~\cite{liang2021swinir}, restoring more details.
	
	
\begin{figure}[t]
		\begin{center}
			\includegraphics[width=0.7\linewidth]{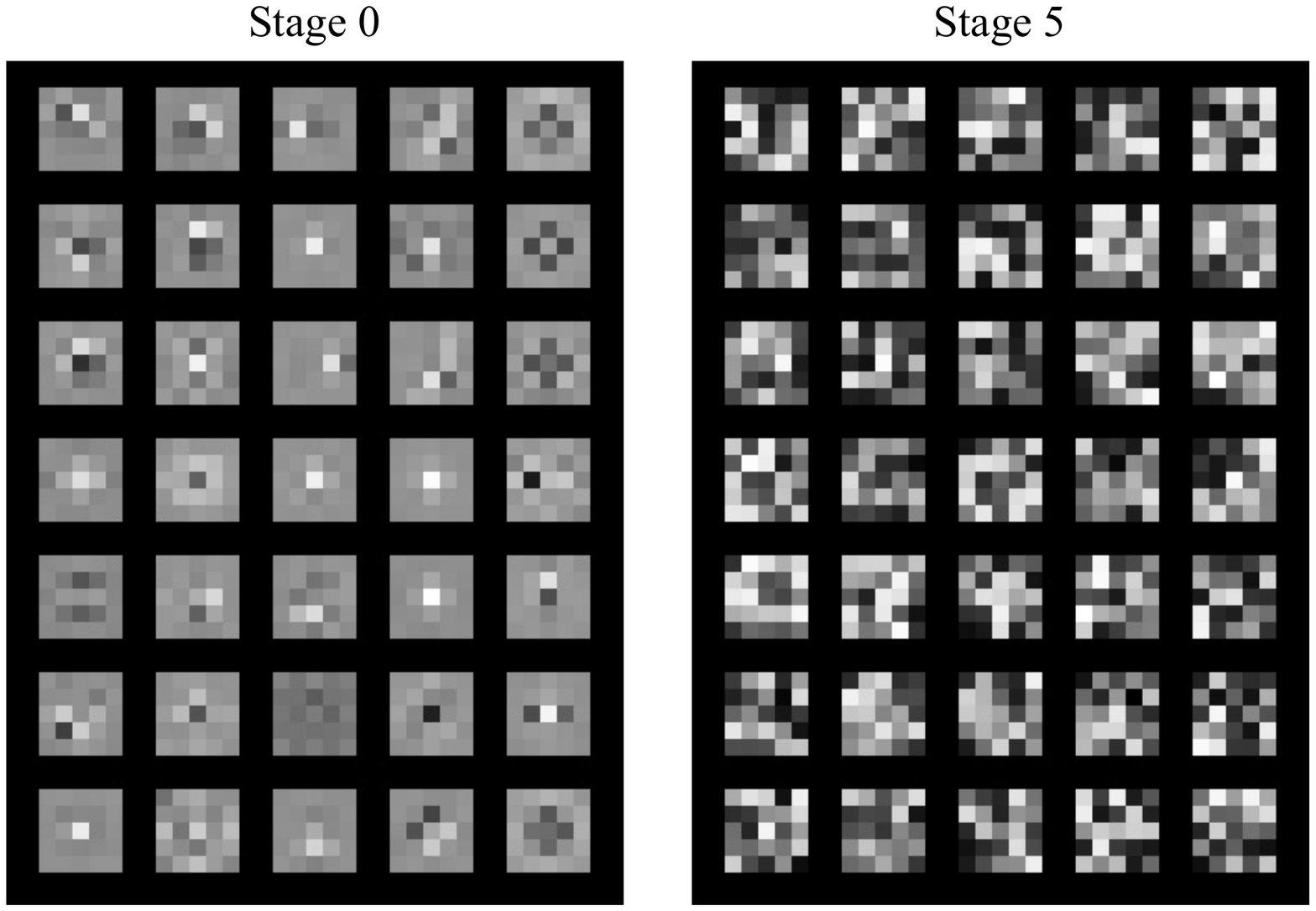}
		\end{center}
		\vspace{-0.2in}
		\caption{Visualization of kernels of the anti-blocking FFN in the first (``Stage 0'') and last (``Stage 5'') Transformer Stage in EDT-B in $\times 2$ SR.}
		\label{fig:kernel}
		\vspace{-0.05in}
	\end{figure}

	%
	%
	\vspace{0.05in}
	\noindent{\textbf{Kernels in Anti-blocking FFN.}} We visualize the kernels of depth-wise convolutions in the Anti-FFN block. As shown in Fig.~\ref{fig:kernel}, we observe a more uniform distribution of kernels at lower layers while diverse representations at higher layers. We find most kernels at lower layers are just like low-pass filters, acting like anti-aliasing filtering to avoid the possible blocking effect caused by window splitting in self-attention, thus meeting our expectations. As for higher layer kernels, they show a high diversity, learning various local contexts. Without the proposed anti-blocking design, there is nearly a 0.1dB drop on both Urban100 and Manga109 in $\times 2$ SR, verifying the necessity of this design.
	
	\section{Datasets}
	
	As illustrated in Sec.~{3.1}, unless specified otherwise, we only use 200K images in ImageNet~\cite{deng2009imagenet} for pre-training. In sec.~{3.5}, we further discuss the effect of data scale on pre-training. As for fine-tuning, datasets of super-resolution (SR), denoising and deraining are described as follows.
	
	\vspace{0.05in}
	\noindent\textbf{Super-Resolution.} The models are trained on 800 images of DIV2K~\cite{agustsson2017ntire} and 2560 images of Flickr2K~\cite{timofte2017ntire}. The evaluation benchmark datasets consist of Set5~\cite{bevilacqua2012low}, Set14~\cite{zeyde2010single}, BSDS100~\cite{martin2001database}, Urban100~\cite{huang2015single} and Manga109~\cite{matsui2017sketch}.
	
	\vspace{0.05in}
	\noindent\textbf{Denoising.} Following~\cite{zhang2020residual,zhang2021plug,liang2021swinir}, we use the combination of 800 DIV2K images, 2650 Flickr2K images, 400 BSD500~\cite{arbelaez2010contour} images and 4744 WED~\cite{ma2016waterloo} images as training data, and use CBSD68~\cite{martin2001database}, Kodak24~\cite{richdodak}, McMaster~\cite{zhang2011color} and Urban100 for evaluation.
	
	\vspace{0.05in}
	\noindent\textbf{Deraining.} Rain100L~\cite{yang2019joint} and Rain100H~\cite{yang2019joint} are used for training and testing. Following~\cite{ren2019progressive}, we exclude 546 training images in Rain100H that have the same background contents with testing images.
	
	
	\section{Training Details}
	
	All experiments are conducted on eight NVIDIA GeForce RTX 2080Ti GPUs, except that multi-task pre-training models and denoising models without downsampling are trained on eight NVIDIA V100 GPUs. All models are trained by the L1 loss.

	\vspace{0.05in}
	\noindent\textbf{Pre-training.} We train models with a batch size (8 GPUs) of 32 per task for 500K iterations. The initial learning rate is set to $2 \times 10^{-4}$ and halved at $[250{\rm K}, 400{\rm K}, 450{\rm K}, 475{\rm K}]$ iterations. We adopt the Adam optimizer with $\beta_{1}=0.9$ and $\beta_{2}=0.99$. The input patch size is set to $48 \times 48$ for SR and $192 \times 192$ for denoising/deraining, thus different tasks share the same feature size $48 \times 48$ in the body part.
	
	\vspace{0.05in}
	\noindent\textbf{Fine-tuning.} We fine-tune the SR models for 500K iterations, denoising models for 800K iterations and deraining models for 200K iterations. The learning rate is $1 \times 10^{-5}$. Other settings remain the same as pre-training.
	
	\vspace{0.05in}
	\noindent\textbf{Training from Scratch.} The initial learning rate is uniformly set to $2 \times 10^{-4}$. We train SR models for 500K iterations, during which the learning rate is halved at $[250{\rm K}, 400{\rm K}, 450{\rm K}, 475{\rm K}]$ iterations. As for denoising, we train models for 800K iterations and the learning rate is halved at $[400{\rm K}, 640{\rm K}, 720{\rm K}, 750{\rm K}]$ iterations. In terms of deraining, we train models for 200K iterations, halving the learning rate at $[100{\rm K}, 150{\rm K}, 180{\rm K}, 190{\rm K}]$ iterations. Other settings are the same as pre-training.
	
	As aforementioned in Sec.~{4.2}, we also train denoising models without downsampling from scratch. Following~\cite{liang2021swinir}, the batch size (8 GPUs) is set to 8 and iteration number is 1600K. The initial learning rate is $2 \times 10^{-4}$ and halved at $[800{\rm K}, 1200{\rm K}, 1400{\rm K}, 1500{\rm K}]$ iterations.
	
	
	
	\begin{table}[t]
		\caption{Quantitative comparison between single-task (``Single'') and multi-related-task (``Multi-Related'') pre-training for EDT-B model on PSNR(dB)/SSIM in SR, denoising and deraining. The multi-task setting of SR includes $\times 2$, $\times 3$ and $\times 4$, denoising includes g15, g25 and g50, and deraining includes light and heavy rain streaks. For clarity, we only provide partial results on several datasets.}
		\renewcommand\arraystretch{1.2}
		\setlength\tabcolsep{4pt}
		\begin{center}
			\begin{tabular}{| c | c | c | c c |}
				\hline
				Task & Dataset & Setting & Single & Multi-Related \\
				\hline
				\multirow{6}{*}{SR} & \multirow{3}{*}{Urban100} & $\times 2$ & 33.95/0.9435 & \textbf{34.27}/\textbf{0.9456} \\
				~ & ~ & $\times 3$ & 29.82/0.8825 & \textbf{30.07}/\textbf{0.8863} \\
				~ & ~ & $\times 4$ & 27.61/0.8275 & \textbf{27.75}/\textbf{0.8317} \\
				\cline{2-5}
				~ & \multirow{3}{*}{Manga109} & $\times 2$ & 40.25/0.9806 & \textbf{40.37}/\textbf{0.9811} \\
				~ & ~ & $\times 3$ & 35.30/0.9542 & \textbf{35.47}/\textbf{0.9550} \\
				~ & ~ & $\times 4$ & 32.22/0.9265 & \textbf{32.39}/\textbf{0.9283} \\
				\hline
				\multirow{6}{*}{Denoising} & \multirow{3}{*}{CBSD68} & g15 & 34.34/0.9348 & \textbf{34.38}/\textbf{0.9352} \\
				~ & ~ & g25 & 31.74/0.8932 & \textbf{31.76}/\textbf{0.8937} \\
				~ & ~ & g50 & 28.56/0.8118 & \textbf{28.57}/\textbf{0.8120} \\
				\cline{2-5}
				~ & \multirow{3}{*}{Urban100} & g15 & 34.94/0.9514 & \textbf{35.04}/\textbf{0.9522} \\
				~ & ~ & g25 & 32.79/0.9298 & \textbf{32.86}/\textbf{0.9307} \\
				~ & ~ & g50 & 29.93/0.8886 & \textbf{29.98}/\textbf{0.8892} \\
				\hline
				\multirow{2}{*}{Deraining} & Rain100L & light & 41.80/0.9900 & \textbf{42.50}/\textbf{0.9905} \\
				\cline{2-5}
				~ & Rain100H & heavy & 33.97/0.9400 & \textbf{34.25}/\textbf{0.9485} \\
				\hline
			\end{tabular}
		\end{center}
		\vspace{-0.1in}
		\label{tab:singlevsmulti}
	\end{table}
	
	\section{Single- and Multi-Task Pre-training}
	\label{sec:singlemulti}
	
\begin{figure}[t]
		\centering
		\begin{minipage}[c]{1.0\linewidth}
			\centering
			\caption{PSNR(dB) comparison between single-task large-batch and multi-task pre-training for EDT-B in $\times 2$ SR. ``Batch'' represents the batch size of a single task.}
			\vspace{0.15in}
			\renewcommand\arraystretch{1.1}
			\setlength\tabcolsep{2pt}
			\begin{tabular}{| c | c | c | c c c c c |}
				\hline
				Type & Tasks & Batch & Set5 & Set14 & BSDS100 & Urban100 & Manga109 \\
				\hline
				Single & $\times 2$ & 32 & 38.56 & 34.71 & 32.57 & 33.95 & 40.25 \\
				Single & $\times 2$ & 96 & 38.61 & \textbf{34.83} & 32.61 & 34.14 & \textbf{40.39} \\
				Multi-Unrelated & $\times 2$, $\times 3$, g15 & 32 & 38.59 & 34.80 & 32.60 & 34.16 & 40.31 \\
				Multi-Related & $\times 2$, $\times 3$, $\times 4$ & 32 & \textbf{38.63} & 34.80 & \textbf{32.62} & \textbf{34.27} & 40.37 \\
				\hline
			\end{tabular}
			\label{tab:largebatch}
		\end{minipage}
		\vspace{0.2in}
		\begin{minipage}[c]{1.0\linewidth}
			\centering
			\includegraphics[width=1.0\textwidth]{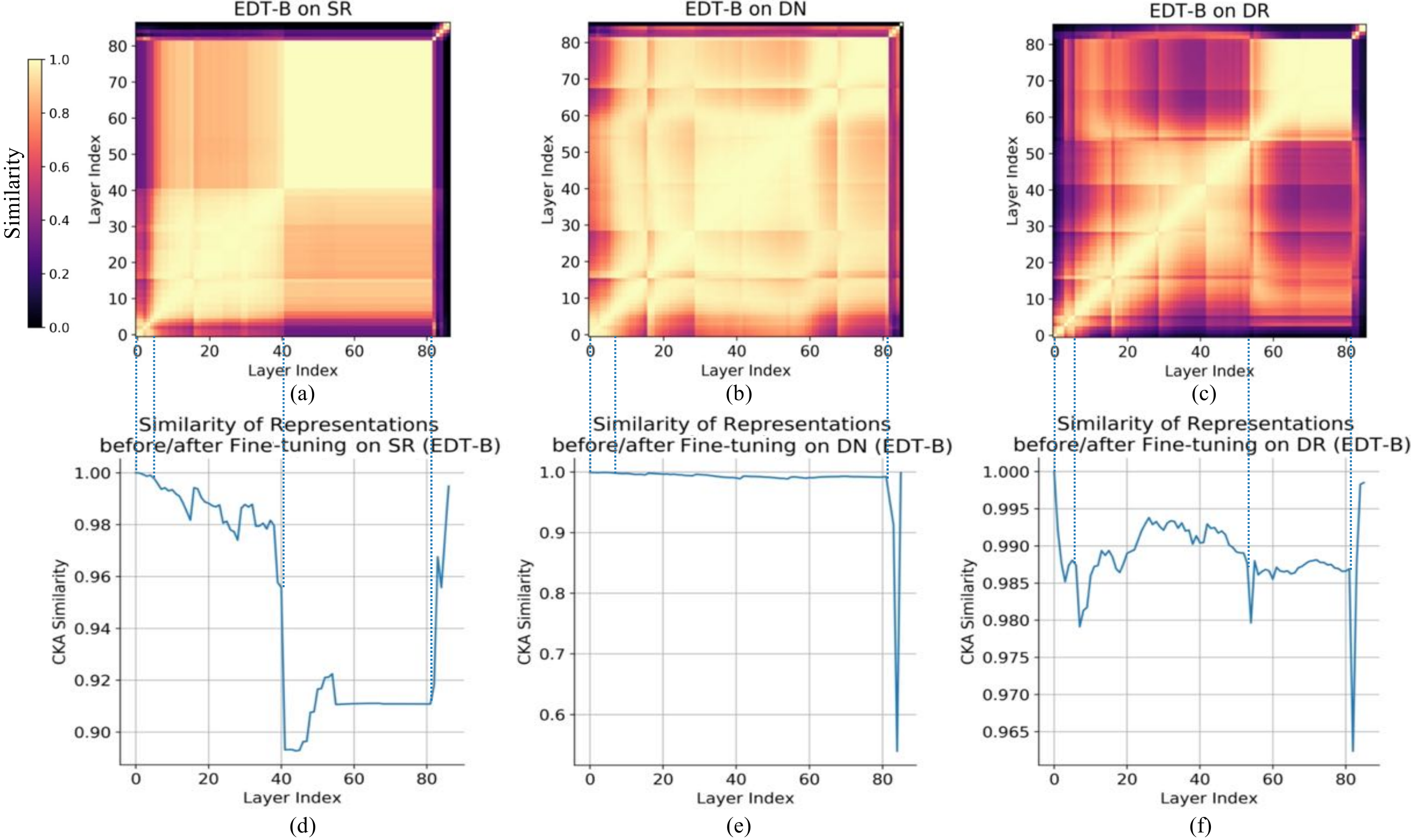}
			\vspace{-0.2in}
			\captionof{figure}{Sub-figures (a)-(c) show CKA similarities between all pairs of layers in $\times 2$ EDT-B SR model, level-15 EDT-B denoising model and light streak EDT-B deraining model with single-task pre-training, and the similarities between before and after fine-tuning are shown in (d)-(f).}
			\label{fig:finetune}
		\end{minipage}
		\vspace{-0.1in}
	\end{figure}

	As shown in Table~\ref{tab:singlevsmulti}, compared to single-task pre-training, multi-related-task setup leads to obviously more improvements on PSNR(dB) and SSIM in all tasks, especially in SR and deraining. Considering the multi-task setting enables the transformer body to see more samples in an iteration, we also conduct a single-task pre-training with a large batch size in Table~\ref{tab:largebatch}. It is observed that multi-related-task pre-training achieves superior or comparable performance, while is efficient in providing initialization for multiple tasks.

	%
	
	
	
	\section{Before and After Fine-tuning}
	
	Considering our models are trained in two stages, pre-training on the ImageNet and fine-tuning on the target dataset, we also study the differences of model representations between before and after fine-tuning. As shown in Fig.~\ref{fig:finetune} (d)-(e), we find fine-tuning mainly changes the higher layer representations of the SR model but has little effect on the denoising model, which shows similar phenomenons to the comparison between without and with pre-training in Sec.~{3.3}. 
	
	As for the deraining task, model representations do not change too much after fine-tuning (the similarities are almost larger than 0.98), where the degrees of changes between the second and the third groups maintain the same level. It is mainly due to that the fine-tuning dataset only contains hundreds of low-resolution images, resulting in limited performance gains. In comparison, as illustrated in Fig.~{2} (f), we observe obvious representation changes between models with and without pre-training, since pre-training helps the model see more data. As a result, our model with pre-training outperforms other methods by a large margin, demonstrating the importance of pre-training.

	\section{Full ImageNet Pre-training}
	
	In Sec.~{3.5}, we show that a larger data scale (from 50K to 400K images) brings more improvements for the single-task pre-training in SR. Here we use the full ImageNet~\cite{deng2009imagenet} dataset  (containing 1.28M images) to explore the upper bound of single-task pre-training. As mentioned in Sec.~{3.5}, we double the pre-training iterations for the data scale of 400K (compared to 200K) so that the data can be fully functional. Thus, we further extend the pre-training period for the full ImageNet setting. From the results in Table~\ref{tab:full_data}, we see that our EDT-B model with full ImageNet pre-training marks a new state of the art on all benchmark datasets. And we suggest that our multi-related-task pre-training strategy would be more promising, which will be studied in the future research.
	
	\begin{table}[t]
		\caption{PSNR(dB) results of different pre-training (single-task) data scales in $\times 2$ SR. ``EDT-B$^\dagger$'' refers to the base model with single-task ($\times 2$ SR) pre-training. The best results are in \textbf{bold}.}
		\setlength\tabcolsep{4pt}
		\begin{center}
			\begin{tabular}{| c | c | c | c  c  c  c  c|}
				\hline
				Model & Data & Iters & Set5 & Set14 & BSDS100 & Urban100 & Manga109 \\
				\hline
				EDT-B & 0 & 500K & 38.45 & 34.57 & 32.52 & 33.80 & 39.93 \\
				\hline
				EDT-B$^\dagger$ & 200K & 500K & 38.56 & 34.71 & 32.57 & 33.95 & 40.25 \\
				EDT-B$^\dagger$ & 400K & 1M & 38.61 & 34.75 & 32.60 & 34.05 & 40.37 \\	
				EDT-B$^\dagger$ & 1.28M & 3M & \textbf{38.66} & \textbf{35.02} & \textbf{32.66} & \textbf{34.58} & \textbf{40.60} \\		
				\hline
			\end{tabular}
		\end{center}
		\vspace{-0.1in}
		\label{tab:full_data}
	\end{table}

	
	\section{Visual Comparison with SOTA Methods}
	
\begin{figure}[h]
		\begin{center}
			\includegraphics[width=1.0\linewidth]{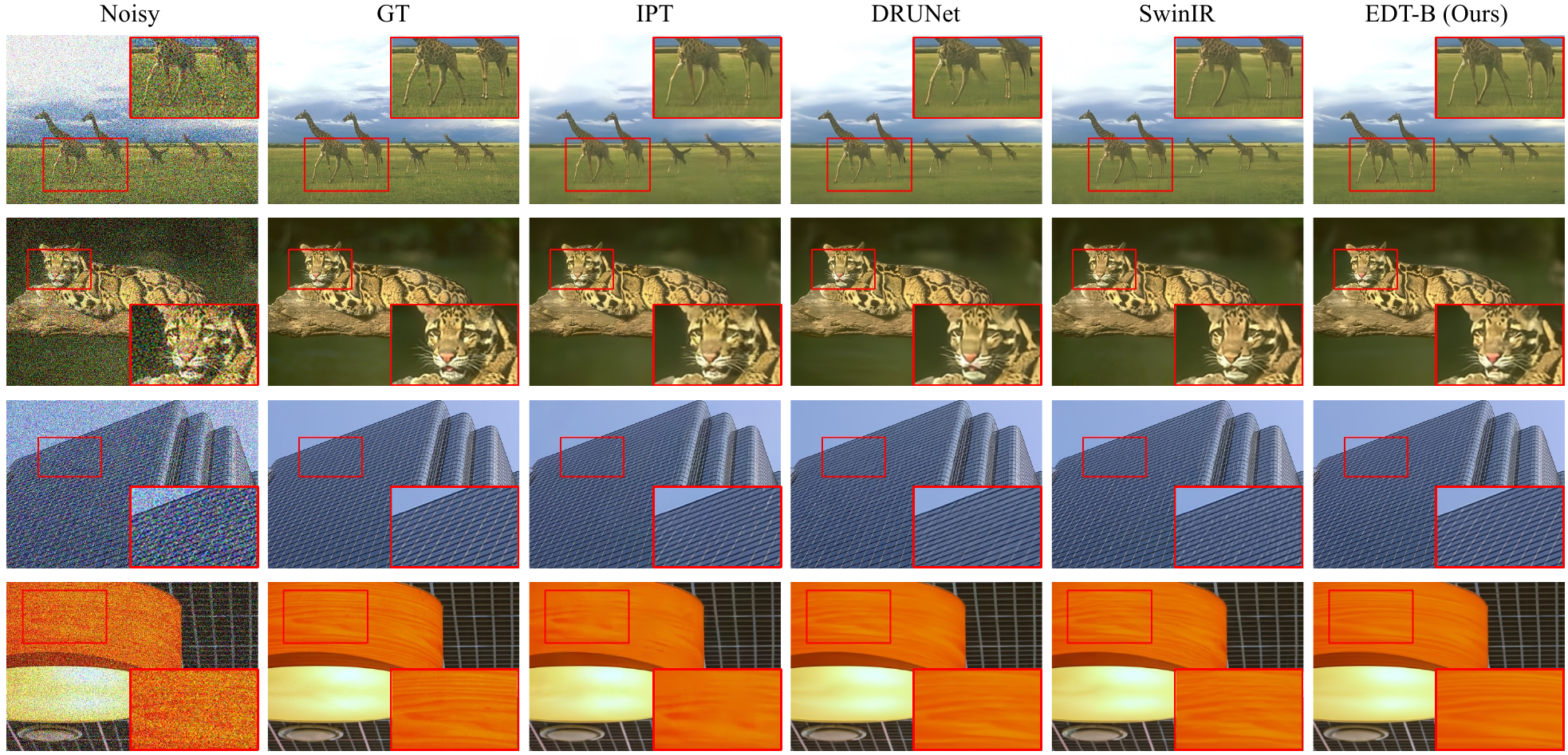}
		\end{center}
		\caption{Qualitative comparison in denoising with noise level 50 on CBSD68~\cite{martin2001database} and Urban100~\cite{huang2015single}.}
		\label{fig:denoise}
	\end{figure}

\begin{figure}[t]
		\begin{center}
			\includegraphics[width=1.0\linewidth]{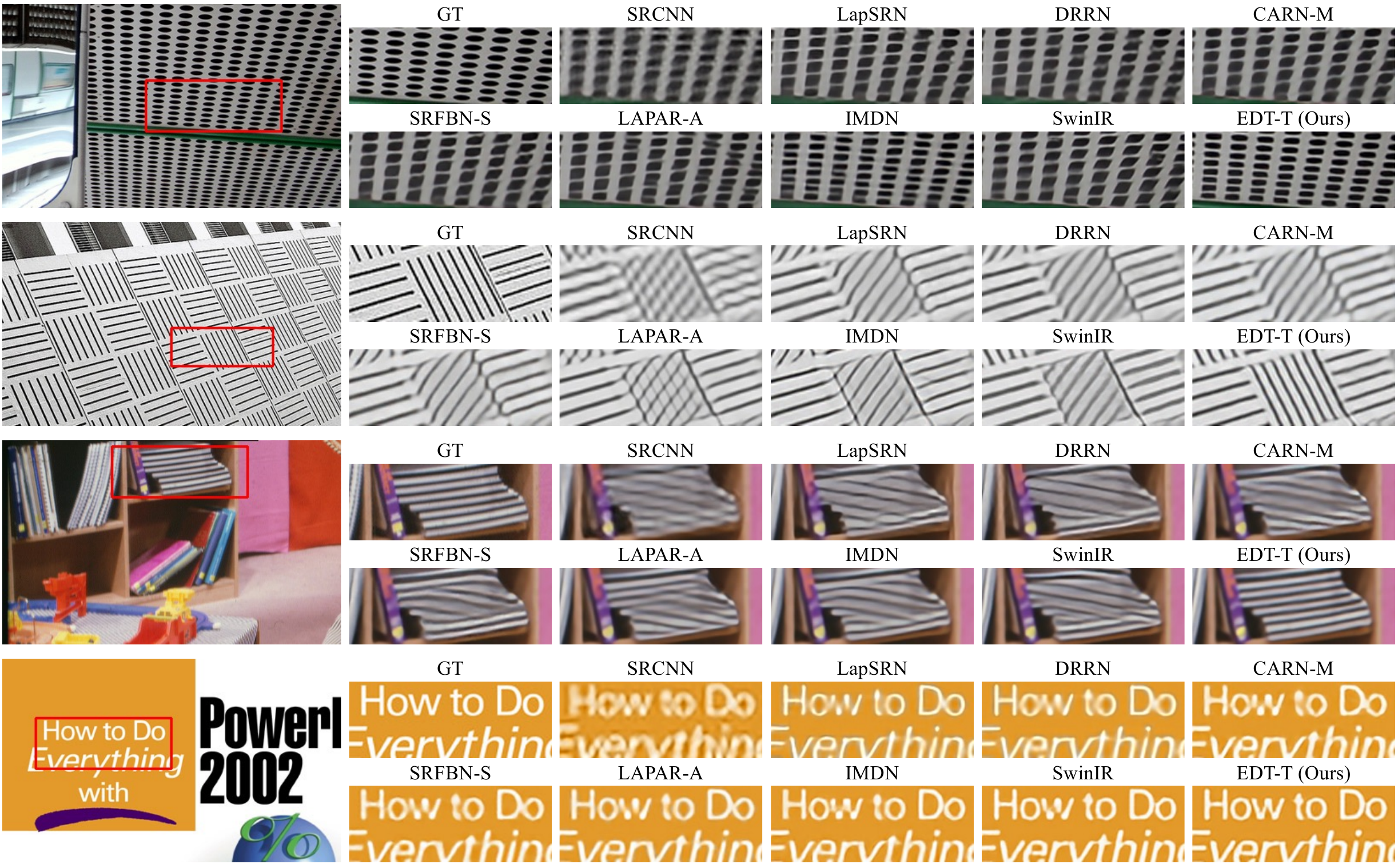}
		\end{center}
		\caption{Qualitative comparison in $\times 4$ lightweight SR on Urban100~\cite{huang2015single} and Set14~\cite{zeyde2010single}.}
		\label{fig:sr}
	\end{figure}

\begin{figure}[t]
		\begin{center}
			\includegraphics[width=1.0\linewidth]{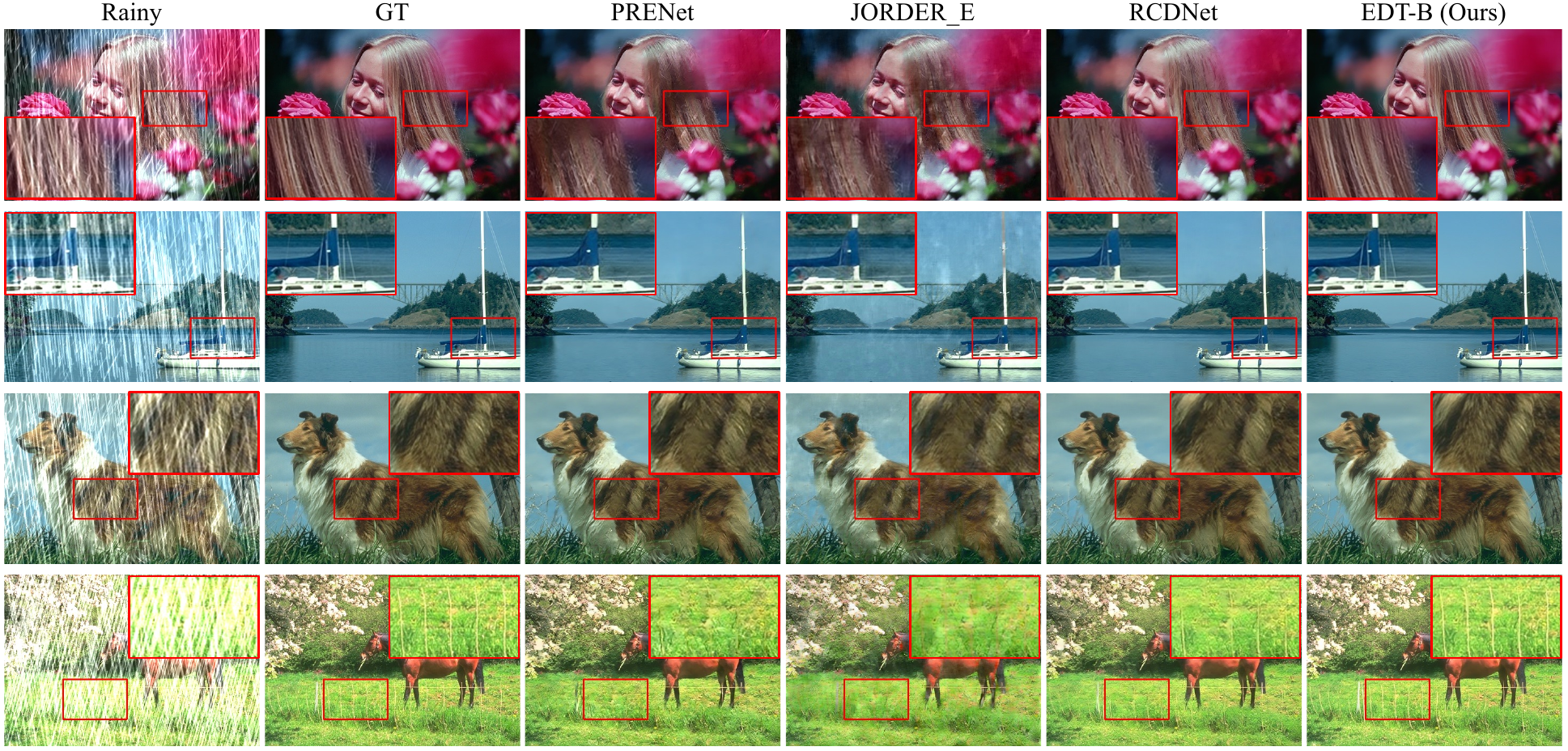}
		\end{center}
		\caption{Qualitative comparison in deraining with heavy rain streaks on Rain100H~\cite{yang2019joint}.}
		\label{fig:derain}
	\end{figure}

	We present more visual examples of denoising, super-resolution and deraining in Fig.~\ref{fig:denoise}, Fig.~\ref{fig:sr} and Fig.~\ref{fig:derain}. Compared with other state-of-the-art methods, the proposed EDT successfully recovers more regular structures and richer textures, producing high-quality images.
	
\end{document}